\documentclass[10pt,twocolumn,letterpaper]{article}

\usepackage[pagenumbers]{cvpr} 

\usepackage{graphicx}
\usepackage{amsmath}
\usepackage{amssymb}
\usepackage{booktabs}
\usepackage{multirow}
\usepackage{array}
\usepackage[accsupp]{axessibility}
\usepackage{listings}

\usepackage[pagebackref,breaklinks,colorlinks]{hyperref}
\usepackage{subcaption}

\usepackage[capitalize]{cleveref}
\crefname{section}{Sec.}{Secs.}
\Crefname{section}{Section}{Sections}
\Crefname{table}{Table}{Tables}
\crefname{table}{Tab.}{Tabs.}
\Crefname{figure}{Figure}{Figures}
\crefname{figure}{Fig.}{Figs.}

\def\cvprPaperID{361} 
\def\confName{CVPR}
\def\confYear{2023}

\usepackage{colortbl}
\usepackage[table,dvipsnames]{xcolor}
\usepackage{xcolor}
\def\todo#1{{\color{purple}{\small\bf\sf#1}}}

\definecolor{tab_yellow}{rgb}{1, 1, 0.7}
\definecolor{tab_orange}{rgb}{1, 0.85, 0.7}
\definecolor{tab_red}{rgb}{1, 0.7, 0.7}

\begin{document}

\title{Balanced Spherical Grid for Egocentric View Synthesis}

\author{Changwoon Choi$^1$, Sang Min Kim$^1$, Young Min Kim$^{1,2}$\\
{\small $^1$Dept. of Electrical and Computer Engineering, Seoul National University, Korea}\\
{\small $^2$Interdisciplinary Program in Artificial Intelligence and INMC, Seoul National University}\\
}

\twocolumn[{
\renewcommand\twocolumn[1][]{#1}
\maketitle
\begin{center}
    \centering
        \captionsetup{type=figure}
        \includegraphics[width=\linewidth]{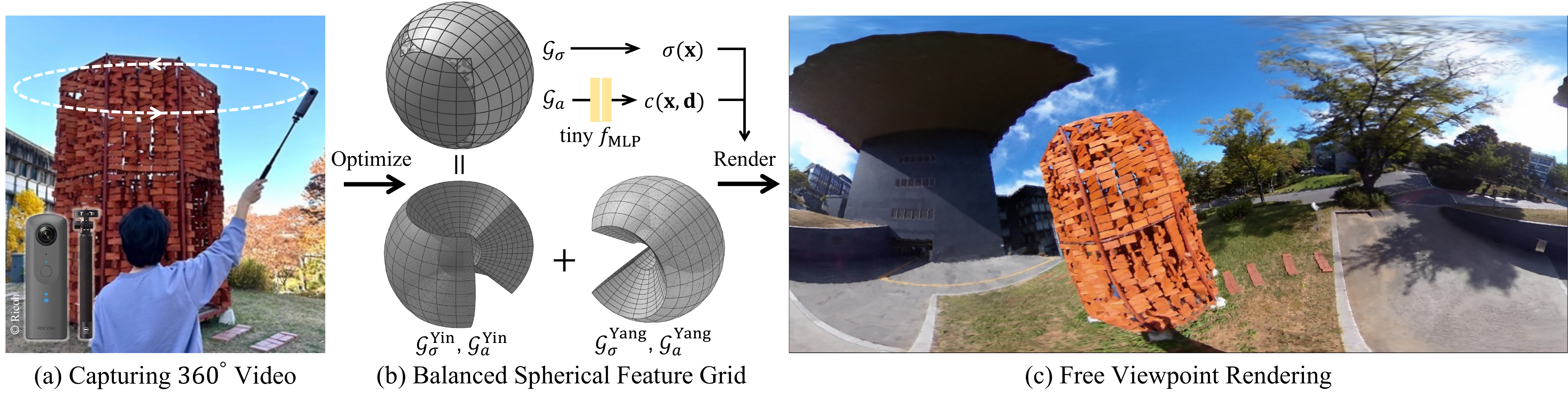}
        \captionof{figure}{We propose a practical solution to reconstruct large-scale scenes from a short egocentric  video. (a) Our scalable capturing setup observes the holistic environment by casually swiping a selfie stick with an omnidirectional camera attached.
        (b) Then we optimize our balanced spherical feature grids which are tailored for the outward-looking setup. 
        (c) EgoNeRF can quickly train and render high-quality images at nearby positions.
        Project page: \url{https://changwoon.info/publications/EgoNeRF}
        } 
        \label{fig:teaser}

\end{center}
}]

\maketitle

\begin{abstract}
\vspace{-0.5em}
We present EgoNeRF, a practical solution to reconstruct large-scale real-world environments for VR assets.
Given a few seconds of casually captured 360 video, EgoNeRF can efficiently build neural radiance fields.
Motivated by the recent acceleration of NeRF using feature grids, we adopt spherical coordinate instead of conventional Cartesian coordinate.
Cartesian feature grid is inefficient to represent large-scale unbounded scenes because it has a spatially uniform resolution, regardless of distance from viewers.
The spherical parameterization better aligns with the rays of egocentric images, and yet enables factorization for performance enhancement.
However, the na\"ive spherical grid suffers from singularities at two poles, and also cannot represent unbounded scenes. 
To avoid singularities near poles, we combine two balanced grids, which results in a quasi-uniform angular grid.
We also partition the radial grid exponentially and place an environment map at infinity to represent unbounded scenes. 
Furthermore, with our resampling technique for grid-based methods, we can increase the number of valid samples to train NeRF volume.
We extensively evaluate our method in our newly introduced synthetic and real-world egocentric 360 video datasets, and it consistently achieves state-of-the-art performance.
\vspace{-0.5em}
\end{abstract}

\section{Introduction}

With the recent advance in VR technology, there exists an increasing need to create immersive virtual environments.
While a synthetic environment can be created by expert designers, various applications also require transferring a real-world environment.
Spherical light fields~\cite{broxton2019low,broxton2020deepview,broxton2020immersive, pozo2019integrated,overbeck2018system} can visualize photorealistic rendering of the real-world environment with the help of dedicated hardware with carefully calibrated multiple cameras.
A few works~\cite{jang2022egocentric,bertel2020omniphotos} also attempt to synthesize novel view images by reconstructing an explicit mesh from an egocentric omnidirectional video.
However, their methods consist of complicated multi-stage pipelines and require pretraining for optical flow and depth estimation networks.

In this paper, we build a system that can visualize a large-scale scene without sophisticated hardware or neural networks trained with general scenes.
We utilize panoramic images, as suggested in spherical light fields.
However, we acquire input with a commodity omnidirectional camera with two fish-eye lenses instead of dedicated hardware.
As shown in~\cref{fig:teaser} (a), the environment can be captured with the omnidirectional camera attached to a selfie stick within less than five seconds.
Then the collected images observe a large-scale scene that surrounds the viewpoints. 
We introduce new synthetic and real-world datasets of omnidirectional videos acquired from both indoor and outdoor scenes.
Combined with Neural Radiance Fields (NeRF)~\cite{mildenhall2021nerf}, the images can train a neural volume that can render fine details or view-dependent effects without explicit 3D models.

\begin{figure}
    \includegraphics[width=\linewidth]{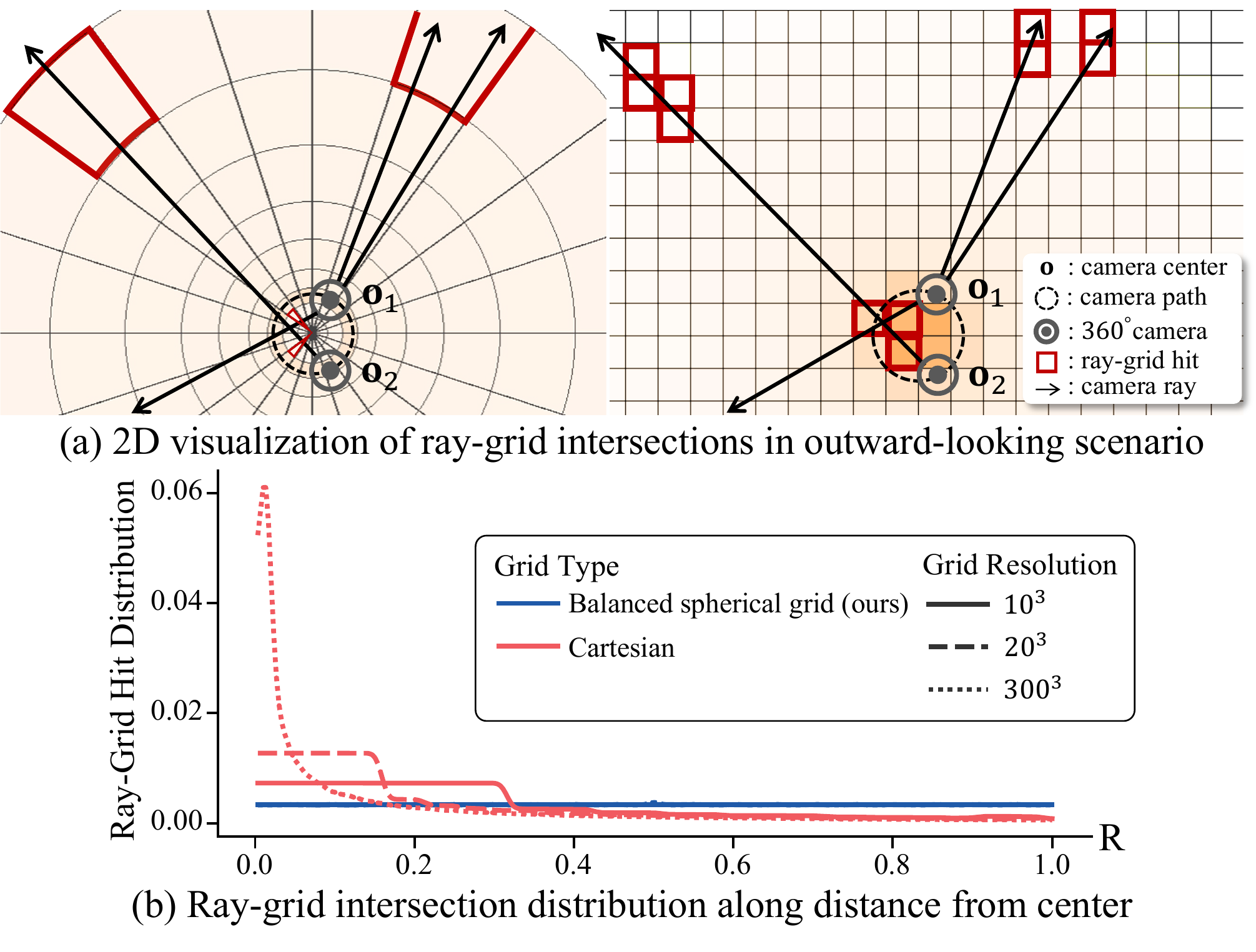}
    \caption{(a) When the camera trajectory is short relative to the scene size, the proposed balanced spherical grid (left) exhibits uniform hitting rate for grid cells whereas the conventional Cartesian grid (right) suffers from non-uniform ray-grid hits. The orange shade indicates the relative density of hit count of the grid cells. (b) Experiments show that spherical coordinates achieve nearly uniform ray-grid hit distribution, while Cartesian coordinate is biased to the center especially when we use a fine-resolution grid.}
    \label{fig:grid_comparison}
\end{figure}

\begin{figure}
    \centering
    \includegraphics[width=\linewidth]{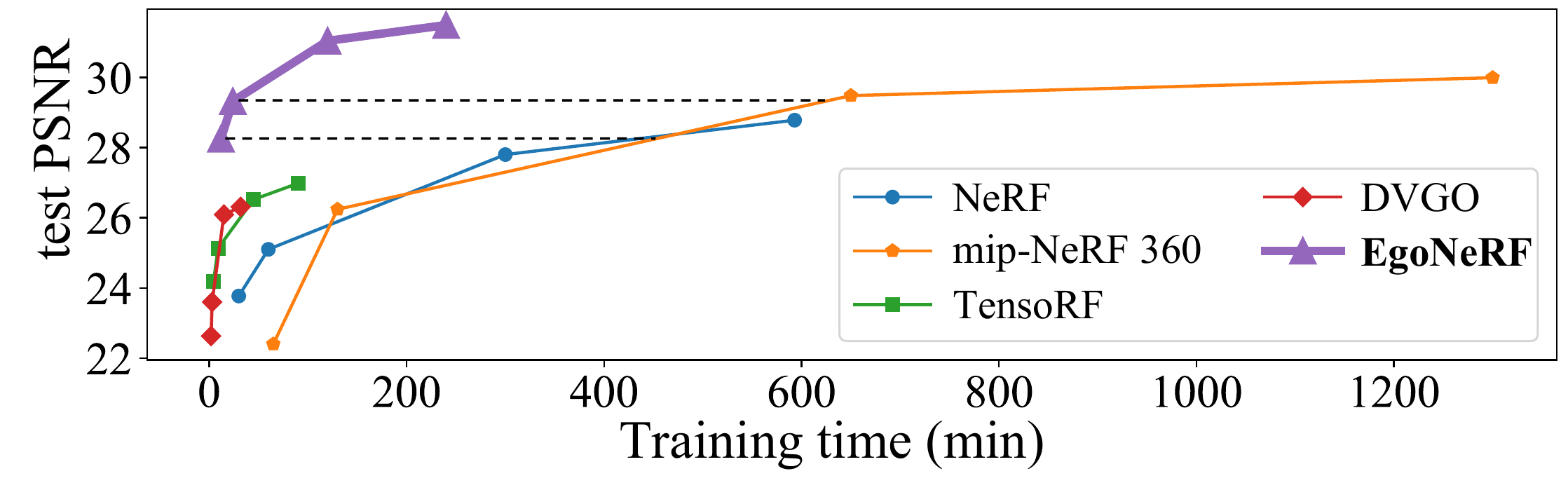}
    \caption{Training curve comparison in \textit{OmniBlender} scenes.}
    \label{fig:time-PSNR}
\end{figure}

To this end, we present Egocentric Neural Radiance Fields, or EgoNeRF, which is the neural volume representation tailored to egocentric omnidirectional visual input.
Although NeRF and its variants with MLP-based methods show remarkable performance in view synthesis, they suffer from lengthy training and rendering time.
The recent Cartesian feature grids can lead to faster convergence~\cite{chen2022tensorf, Sun_2022_CVPR} for rendering a bounded scene with an isolated object, but they have several limitations for our datasets which mostly contain inside-out views of large scenes:
(1) The uniform grid size, regardless of distance from the camera, is insufficient to represent fine details of near objects and extravagant for coarse integrated information from far objects.
(2) Cartesian grid suffers from non-uniform ray-grid hits in the egocentric scenario as demonstrated in~\cref{fig:grid_comparison}, thus, as pointed in~\cite{Sun_2022_CVPR}, prior arts need careful training strategies such as progressive scaling~\cite{chen2022tensorf,Sun_2022_CVPR} or view-count-adaptive per-voxel learning rate~\cite{Sun_2022_CVPR}.
EgoNeRF models the volume using a spherical coordinate system to cope with the aforementioned limitations.
\Cref{fig:time-PSNR} shows that EgoNeRF converges faster compared to MLP-based methods (NeRF~\cite{mildenhall2021nerf} and mip-NeRF 360~\cite{Barron_2022_CVPR}) and has higher performance compared to Cartesian grid methods (TensoRF~\cite{chen2022tensorf} and DVGO~\cite{Sun_2022_CVPR}).

Our spherical grid is designed to be balanced in any direction, which leads to a more efficient data structure for the large-scale environment.
The na\"ive spherical grid contains high valence vertices at two poles, and, when adapted as a feature grid for neural volume rendering, the polar regions suffer from undesirable artifacts.
We exploit a quasi-uniform angular grid by combining two spherical grids~\cite{kageyama2004yin}.
In the radial direction, the grid intervals increase exponentially, which not only allows our representation to cover large spaces but also makes the spherical frustum have a similar length in the angular and radial directions.
We add an environment map at infinite depth, which is especially useful for outdoor environments with distant backgrounds such as skies.
Last but not least, we propose an efficient hierarchical sampling method exploiting our density feature grid without maintaining an additional coarse density grid.

We demonstrate that our proposed approach can lead to faster convergence and high-quality rendering with a small memory footprint in various scenarios for large-scale environments.
EgoNeRF is expected to create a virtual rendering of large scenes from data captured by non-expert users, which cannot be easily modeled with 3D assets or conventional NeRF.

\section{Related Works}

\paragraph{Visualizing Omnidirectional View of Scenes}
Panoramic images are widely used in many applications for remote experiences. 
After captured by photo-stitching apps or dedicated hardware, they allow users to rotate around the captured position.
However, we need additional information to allow the full 6 DoF movement in the scene.
Prior works propose sophisticated camera rigs to capture spherical light fields~\cite{broxton2019low, broxton2020deepview, broxton2020immersive, pozo2019integrated, overbeck2018system}.
Given multi-view images, they enable synthesizing images at novel viewpoints by reconstructing 3D mesh or multi-sphere images instead of multi-plane images in ordinary images.
With additional depth information, recent works demonstrate novel view synthesis with a single panoramic image~\cite{hara2022enhancement, hsu2021moving}.
The depth channel is acquired from RGBD camera or approximated coarse planar facades.

In contrast, we assume more casual input, using commodity 360$^\circ$ camera with two fish-eye lenses to capture a short video clip of the large-scale scene.
A few works also explored the same setup~\cite{bertel2020omniphotos, jang2022egocentric} and represented the scene with a deformed proxy mesh with texture maps using pre-trained neural networks for optical flow and depth estimation.
Our pipeline is simpler as we train a neural network with the captured sequence of images without any pre-trained network.
We combine the visualization pipeline for large-scale scenes with NeRF formulation and can capture complex view-dependent effects and fine structures, unlike reconstructed textured mesh.

\paragraph{Practical Variants of NeRF}
NeRF~\cite{mildenhall2021nerf} flourished in the field of novel view synthesis, showing photorealistic quality with its simple formulation.
However, the original NeRF formulation exhibits clear drawbacks, such as lengthy training and rendering time, and the difficulty of deformation or scene edits.
Many follow-up works exploded, overcoming the limitations in various aspects~\cite{Barron_2021_ICCV, Barron_2022_CVPR, Martin-Brualla_2021_CVPR, Niemeyer_2021_CVPR, Park_2021_ICCV, Srinivasan_2021_CVPR, choi2022ibl}. 
Here we specifically focus on practical extension for fast rendering and training.
NeRF represents a scene as a single MLP that maps coordinates into color and volume density. 
It is slow in rendering and optimization as the volume rendering requires multiple forward passes of the MLP.

To accelerate the rendering speed, radiance is represented with an explicit voxel grid storing features~\cite{liu2020neural, yu2021plenoctrees, hedman2021baking}.
However, they train the network by distilling information from pre-trained NeRF, which even lengthens the training time.
More recent works exploit various data structures to directly optimize the feature grid~\cite{chen2022tensorf, Sun_2022_CVPR, muller2022instant, Fridovich-Keil_2022_CVPR}.
They have shown that employing an explicit feature grid achieves fast optimization without sacrificing quality.
The feature grids are defined on the Cartesian coordinate system, which assumes a scene within a bounding box.
These are not suitable for representing large-scale scenes whose viewpoints observe outside of the captured locations.

The na\"ive strategy to choose ray samples wastes most samples and it leads to slow convergence since many regions are either free spaces or occluded by other objects in the real world.
To increase the sample efficiency, the original NeRF~\cite{mildenhall2021nerf} employs a hierarchical sampling strategy for the volume density and maintains two density MLPs for coarse and fine resolution, respectively.
In the same context, M\"uller et al.~\cite{muller2022instant} maintain additional multi-scale occupancy grids to skip ray marching steps.
Hu et al.~\cite{Hu_2022_CVPR} allocate dense momentum voxels for valid sampling, and Sun et al.~\cite{Sun_2022_CVPR} also use an extra coarse density voxel grid.
Maintaining separate coarse feature grids or neural networks requires additional memory and increases computational burdens. 
We propose an efficient sampling strategy and quickly train a volume that represents a large-scale environment.
\section{Feature Grid Representation for EgoNeRF}
\label{sec:feature_grid}

\begin{figure*}
    \centering
    \includegraphics[width=\linewidth]{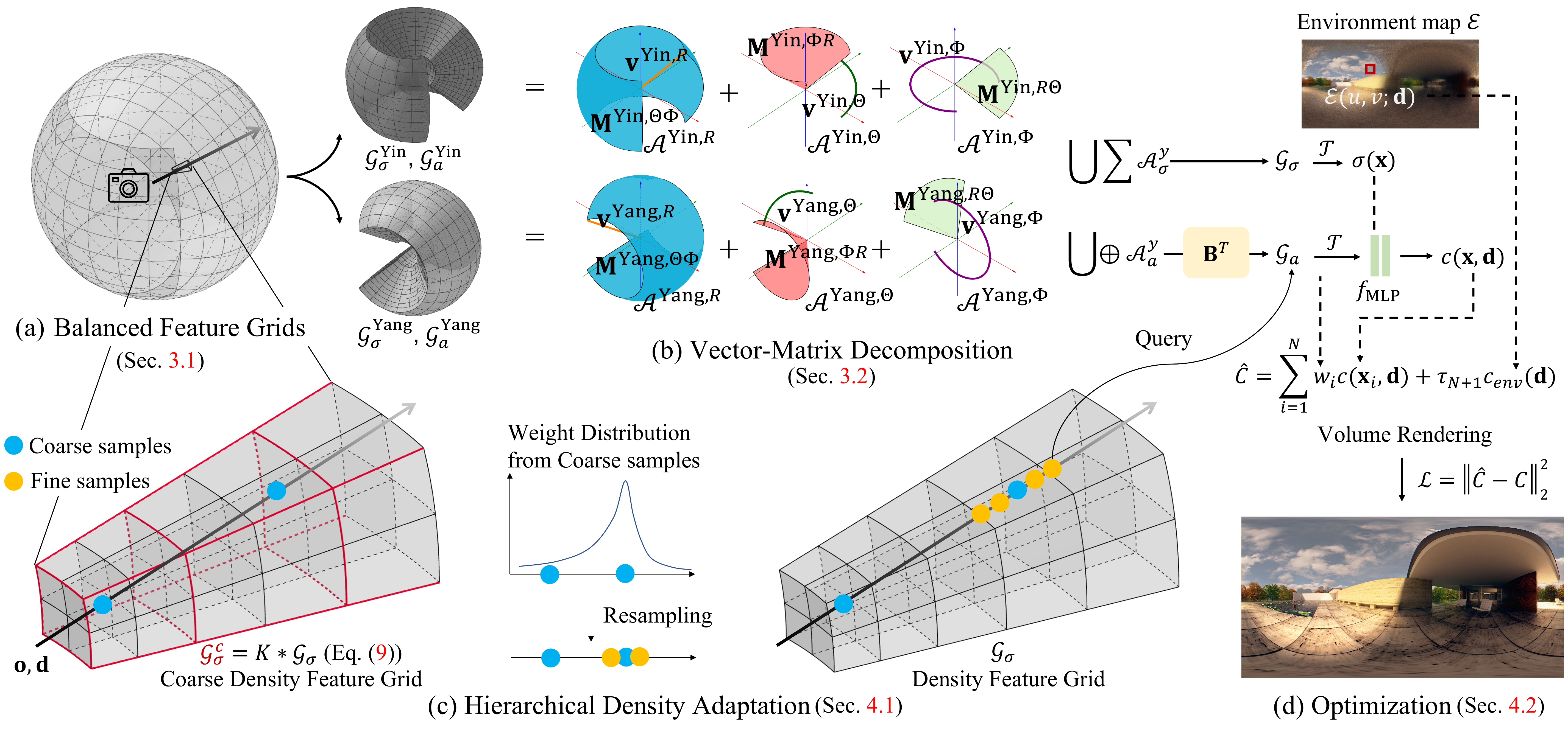}
    \caption{Overview of our method. (a) We represent radiance fields as features stored in balanced feature grids $\mathcal{G}_\sigma\text{, } \mathcal{G}_a$, (b) which are further decomposed into vector and matrix components. (c) The hierarchical sampling is conducted by obtaining a coarse density grid from the density feature grid on the fly during optimization. (d) The balanced feature grids are optimized with photometric loss.}
    \label{fig:method_overview}
\end{figure*}

EgoNeRF utilizes feature grids to accelerate the neural volume rendering of NeRF.
Feature grids in previous works employ a Cartesian coordinate system, which regularly partition the volume in $xyz$ axis~\cite{liu2020neural, yu2021plenoctrees, hedman2021baking}.
To better express the egocentric views captured from omnidirectional videos, we use a spherical coordinate system.
We modify the spherical coordinate in both angular and radial partitions to efficiently express outward views of the surrounding environment, as described in \cref{subsec:grid_description}.
For rendering and training, the values are interpolated from the feature grid, which can be further factorized to reduce the memory and accelerate the learning~\cite{chen2022tensorf} (\cref{subsec:grid_NeRF}).
With our balanced feature grid, individual cells produce a uniform hitting rate of rays.

\subsection{Balanced Spherical Grid}
\label{subsec:grid_description}

Our balanced spherical grid is composed of the angular partition and the radial partition.
\paragraph{Angular Partitions}
The desirable angular partition should result in regular shapes and be easily parameterized.
When we regularly partition on the angle parameters, the na\"ive spherical coordinate system results in irregular grid partitions, which severely distort the two polar regions.
Existing regular partitions do not maintain orthogonal axis parameterization~\cite{greger1998irradiance}, which hinders further factorization.

As a simple resolution, we only use the quasi-uniform half of the ordinary spherical coordinate system and combine two of them~\cite{kageyama2004yin}.
The two grids are referred to as the Yin grid and Yang grid, respectively, which have identical shapes and sizes as shown in \cref{fig:teaser} (b) and \cref{fig:method_overview} (a).
Together they can cover the entire sphere with minimal overlap, similar to the two regions of a tennis ball.

The Yin grid is defined as:
\begin{equation}
    (\pi/4 \leq \theta \leq 3\pi/4) \cap (-3\pi/4 \leq \phi \leq 3\pi/4),
\end{equation}
where $\theta$ is colatitude and $\phi$ is longitude.
The axis of another component grid, namely the Yang grid, is located at the equator of the Yin grid:
\begin{equation}
    \begin{bmatrix}x^{\text{Yin}}\\ y^{\text{Yin}}\\ z^{\text{Yin}}\end{bmatrix} = M \begin{bmatrix}x^{\text{Yang}}\\ y^{\text{Yang}}\\ z^{\text{Yang}}\end{bmatrix},
    M =
    \begin{bmatrix}
        -1 & 0 & 0\\
        0 & 0 & 1\\
        0 & 1 & 0
    \end{bmatrix}.
\end{equation}
We discretize the angular grid of Yin and Yang grid by $N_\theta^y$ and $N_\phi^y$ partitions for $\theta^y, \phi^y$ axis respectively, where $y\in\{\text{Yin}, \text{Yang}\}$.
The partition is uniform in angles leading to the grid size of
\begin{equation}
    \Delta \theta^y = {\pi\over 2} {1\over N_\theta^y}, \; \Delta \phi^y = {3\pi\over 2} {1\over N_\phi^y}.
\end{equation}

\paragraph{Radial Partitions}
By adopting the spherical coordinate system, the grid cells cover larger regions as $r$ increases.
This is desired in the egocentric setup, as the panoramic image capture more detailed close-by views of central objects while distant objects occupy a small area on the projected images.
We further make the grid along the $r$ axis increase exponentially for far regions such that the resulting cell exhibit similar lengths in the angular and radial direction.

Specifically, if we denote the radial scales of both the Yin and Yang grids as $r^y$,
\begin{equation}
    r_i^y = r_0 k^{i-1}, \; R_{\text{max}} = r_0 k^{N_r^y - 1},
\end{equation}
where $R_{\text{max}}$ is the radius of the scene bounding sphere and constant value $r_0$ is the radius of the first spherical shell.
We set the grid interval to $r_0$ for the grid interval less than $r_0$.

We can optionally use the environment map for outdoor or large indoor environments. 
Our spherical grid is still bounded by $R_\text{max}$, limiting the size of the environment.
The environment map denoted as $\mathcal{E} \in \mathbb{R}^{H\times W \times 3}$, is a simple equirectangular image and represents what is visible at an almost infinite distance.

\subsection{Feature Grid as Radiance Field}
\label{subsec:grid_NeRF}
Now we describe our radiance field representation with the balanced spherical feature grid.
Given a set of egocentric images with corresponding camera parameters, EgoNeRF aims to reconstruct 3D scene representation and synthesize novel view images.
Instead of regressing for the volume density $\sigma$ and color $c$ from MLP~\cite{mildenhall2021nerf}, we build explicit feature grids of the density $\mathcal{G}_\sigma$ and the appearance $\mathcal{G}_a$ which serve as the mapping function.
Both grids are composed of our balanced spherical grids of resolution $2N_r^y\times N_\theta^y \times N_\phi^y$, as defined in~\cref{subsec:grid_description}.
The density grid $\mathcal{G}_\sigma \in \mathbb{R}^{2N_r^y\times N_\theta^y \times N_\phi^y}$ has a single channel which stores the explicit volume density value, and the appearance grid $\mathcal{G}_a \in \mathbb{R}^{2N_r^y \times N_\theta^y \times N_\phi^y \times C}$ stores $C$-dimensional neural appearance features.
The volume density and color at position $\mathbf{x}$ and viewing direction $\mathbf{d}$ are obtained by
\begin{equation}
    \sigma(\mathbf{x}) = \mathcal{T}(\mathcal{G}_\sigma, \mathbf{x}), \, c(\mathbf{x}, \mathbf{d}) = f_{\text{MLP}}(\mathcal{T}(\mathcal{G}_a , \mathbf{x}), \mathbf{d}),
    \label{eq:querying}
\end{equation}
where $\mathcal{T}$ denotes a trilinear interpolation, and $f_{\text{MLP}}$ is a tiny MLP that decodes the neural feature to color.

Inspired by~\cite{chen2022tensorf}, we further decompose the feature tensor into vectors and matrices as shown in \cref{fig:method_overview} (b):
\newcommand{\seq}{\,{=}\!} 
\newcommand{\sotimes}{\,{\otimes}\,} 
\newcommand{\ssplus}{\!{+}\,} 
\newcommand{\splus}{\,{+}\,}
\small
\begin{align}
    \mathcal{G}_\sigma^y &\seq \sum_{n=1}^{N_\sigma} \mathbf{v}_{\sigma, n}^{y, R} \sotimes \mathbf{M}_{\sigma, n}^{y, \Theta \Phi} \ssplus \mathbf{v}_{\sigma, n}^{y, \Theta} \sotimes \mathbf{M}_{\sigma, n}^{y, \Phi R} \ssplus \mathbf{v}_{\sigma, n}^{y, \Phi} \sotimes \mathbf{M}_{\sigma, n}^{y, R \Theta} \nonumber \\
    &\seq \sum_{n=1}^{N_\sigma} \sum_{m\in{R\Theta\Phi}} \mathcal{A}_{\sigma, n}^{y, m}\text{,}
\end{align}
\begin{equation}
    \mathcal{G}_a^y  \seq \sum_{n=1}^{N_a} \mathcal{A}_{a, n}^{y, R}\sotimes \mathbf{b}_{3n-2}^y \splus \mathcal{A}_{a, n}^{y, \Theta}\sotimes \mathbf{b}_{3n-1}^y \splus \mathcal{A}_{a, n}^{y, \Phi}\sotimes \mathbf{b}_{3n}^y\text{,}
\end{equation}
\begin{equation}
    \mathcal{G}_\sigma = \bigcup\limits_{y\in Y} \mathcal{G}_\sigma^y, \mathcal{G}_a = \bigcup\limits_{y\in Y} \mathcal{G}_a^y, Y = \{\text{Yin, Yang}\}\text{,}
\end{equation}
\normalsize
where $\otimes$ represents the outer product and  $\mathbf{v}, \mathbf{b}, \mathbf{M}$ represents vector and matrix factors.
This low-rank tensor factorization significantly reduces the space complexity from $\mathcal{O}(n^3)$ to $\mathcal{O}(n^2)$.
With the minimal overhead of storing two grids, we can maintain regular angular components and yet factorize the grid using spherical parameterization.
The full decomposed formulation is described in the supplementary material.
\section{Training EgoNeRF}
\label{sec:EgoNeRF}

We utilize the balanced spherical grids to represent the volume density $\sigma$ and color $c$, which are stored in $\mathcal{G}_\sigma$ and $\mathcal{G}_a$, respectively.
In this chapter, we describe the technical details of the optimization process of our proposed method.

\subsection{Hierarchical Density Adaptation}
\label{subsec:importance_sampling}

As the scenes typically contain sparse occupied regions, we adapt the hierarchical sampling strategy of the original NeRF~\cite{mildenhall2021nerf} for feature grids.
While other recent variants using feature grid~\cite{muller2022instant,Hu_2022_CVPR,Sun_2022_CVPR} maintain a dedicated data structure for the coarse grid, we exploit our dense geometry feature grid $\mathcal{G}_\sigma$ for the first coarse sampling stage without allocating additional memory for the coarse grid.

The hierarchical sampling strategy first samples coarse $N_c$ points along the ray to obtain a density estimate $\sigma$ from which we can sample fine $N_f$ points with importance sampling.
However, evaluating $\sigma$ with dense $\mathcal{G}_\sigma$ at the coarsely sampled points might skip the important surface regions.
Therefore, we obtain $\sigma$ value from a coarser density feature grid which can be obtained on the fly by applying a non-learnable convolution kernel $K$:
\begin{equation}
    \sigma(\mathbf{x_{\text{coarse}}}) = \mathcal{T}(\mathcal{G}_\sigma^{c}, \mathbf{x}_{\text{coarse}}) = \mathcal{T}(K * \mathcal{G}_\sigma, \mathbf{x}_{\text{coarse}}).
\end{equation}
We use the average pooling kernel as $K$.
It is reasonable to define a coarse grid by convolving the dense grid because our density grid $\mathcal{G}_\sigma$ stores the volume density itself, which has a physical meaning, not neural features.

From the volume density values of coarsely sampled points, we calculate weights for importance sampling by
\begin{equation}
    w_i = \tau_i (1 - e^{-\sigma_i \delta_i}), i\in [1, N_c],
\end{equation}
where $\delta_i$ is the distance between coarse samples, $\tau_i =\  e^{-\sum_{j=1}^{i-1}{\sigma_j\delta_j}}$ is the accumulated transmittance.
Then the fine $N_f$ locations are sampled from the filtered probability distribution. 
Finally, the volume density $\sigma$ and color $c$ at $N_c + N_f$ samples are used to render pixels.

\subsection{Optimization}
\label{subsec:training}

The images of EgoNeRF are synthesized by applying the volume rendering equation along the camera ray~\cite{mildenhall2021nerf} and the optional environment map.
Specifically, the points $\mathbf{x}_i = \mathbf{o} + t_i\mathbf{d}$ along the camera ray from camera position $\mathbf{o}$ and ray direction $\mathbf{d}$ are accumulated to find the pixel value by
\begin{equation}
    \hat{C}=\sum_{i=1}^{N}\tau_i (1-e^{-\sigma(\mathbf{x}_i) \delta_i})c(\mathbf{x}_i,\mathbf{d}) + \tau_{N+1}c_{\text{env}}(\mathbf{d}).
    \label{eq:rendering}
\end{equation}
$N=N_c + N_f$ is the number of samples as described in~\cref{subsec:importance_sampling}.
$\sigma(\mathbf{x})$ and $c(\mathbf{x}, \mathbf{d})$ are obtained from our balanced feature grids in~\cref{eq:querying}.
Since the size of our feature grid is exponentially increasing along the $r$ direction, we distribute $N_c$ coarse samples exponentially rather than uniformly.
The second term in~\cref{eq:rendering} is fetched from the environment map
\begin{equation}
    c_{\text{env}}(\mathbf{d}) = \mathcal{E}(u, v; \mathbf{d}),
\end{equation}
where the sampling position $(u,v)$ is only dependent on the viewing direction $\mathbf{d}$.
The effect of the environment map is further discussed in~\cref{subsec:ablation}.

Finally, we optimize the photometric loss between rendered images and training images
\begin{equation}
    \mathcal{L} = \frac{1}{|\mathcal{R}|}\sum_{\mathbf{r} \in \mathcal{R}} \left\lVert \hat{C}(\mathbf{r}) - C(\mathbf{r}) \right\rVert_2^2,
\end{equation}
where $\mathcal{R}$ is a randomly sampled ray batch, $\hat{C}(\mathbf{r}), C(\mathbf{r})$ are rendered and the ground-truth color of the pixel corresponding to ray $\mathbf{r}$.
With the simple photometric loss, our feature grids $\mathcal{G}_\sigma, \mathcal{G}_a$, decoding MLP $f_{\text{MLP}}$, and environment map $\mathcal{E}$ are jointly optimized.
For real-world datasets, in which camera poses are not perfect, we additionally optimize a TV loss~\cite{rudin1994total} at our feature grid to reduce noise.
Furthermore, since our balanced feature grid guarantees a nearly uniform ray-grid hitting rate, EgoNeRF does not need a coarse-to-fine reconstruction approach for robust optimization used in other feature grid-based methods~\cite{Sun_2022_CVPR, chen2022tensorf}.

\begin{table*}

\centering
\resizebox{0.98\linewidth}{!}{
\begin{tabular}{@{}l@{\:}l| ccccc | ccccc| ccccc}
\toprule
\multirow{3}{*}{Step} & \multirow{3}{*}{Method} & \multicolumn{10}{c|}{\textit{OmniBlender}} & \multicolumn{5}{c}{\textit{Ricoh360}}\\
& & \multicolumn{5}{c|}{Indoor} & \multicolumn{5}{c|}{Outdoor} & \\
 & & PSNR & $\text{PSNR}^{\text{WS}}$ & LPIPS& SSIM & $\text{SSIM}^{\text{WS}}$ & PSNR &  $\text{PSNR}^{\text{WS}}$& LPIPS & SSIM & $\text{SSIM}^{\text{WS}}$ & PSNR &  $\text{PSNR}^{\text{WS}}$& LPIPS & SSIM & $\text{SSIM}^{\text{WS}}$\\
 
 \midrule
 
 \multirow{5}{*}{5k} & NeRF~\cite{mildenhall2021nerf} &\cellcolor{tab_orange}26.25 &\cellcolor{tab_orange}27.27 &\cellcolor{tab_orange}0.500 &\cellcolor{tab_orange}0.726 &\cellcolor{tab_orange}0.710 &\cellcolor{tab_yellow}22.36 &\cellcolor{tab_yellow}23.62 &\cellcolor{tab_yellow}0.524 &\cellcolor{tab_yellow}0.651 &\cellcolor{tab_yellow}0.611 & 22.09 & 23.82 & 0.576 & 0.649 & 0.623\\\
 & mip-NeRF 360~\cite{Barron_2022_CVPR} & 23.51 & 24.41 & 0.628 & 0.649 & 0.613 & 21.76 & 23.03 & 0.545 & 0.614 & 0.568 & 22.30 & 24.12 &\cellcolor{tab_yellow}0.555 & 0.632 & 0.604\\
 & TensoRF~\cite{chen2022tensorf} &\cellcolor{tab_yellow}25.91 &\cellcolor{tab_yellow}26.93 &\cellcolor{tab_yellow}0.553 &\cellcolor{tab_yellow}0.722 &\cellcolor{tab_yellow}0.708 &\cellcolor{tab_orange}23.21 &\cellcolor{tab_orange}24.74 &\cellcolor{tab_orange}0.500 &\cellcolor{tab_orange}0.672 &\cellcolor{tab_orange}0.645 & \cellcolor{tab_orange}23.20 &\cellcolor{tab_orange}25.16 &\cellcolor{tab_orange}0.542 &\cellcolor{tab_orange}0.676 &\cellcolor{tab_orange}0.658\\
 & DVGO~\cite{Sun_2022_CVPR} & 24.26 & 25.29 & 0.633 & 0.689 & 0.666 & 21.70 & 23.15 & 0.570 & 0.642 & 0.605 &\cellcolor{tab_yellow}22.45 &\cellcolor{tab_yellow}24.59 & 0.573 &\cellcolor{tab_yellow}0.664 &\cellcolor{tab_yellow}0.646 \\
 & EgoNeRF & \cellcolor{tab_red}28.87 & \cellcolor{tab_red}30.06 & \cellcolor{tab_red}0.310 & \cellcolor{tab_red}0.803 &\cellcolor{tab_red}0.803 &\cellcolor{tab_red}27.90 & \cellcolor{tab_red}29.31 & \cellcolor{tab_red}0.167 & \cellcolor{tab_red}0.844 & \cellcolor{tab_red}0.832 &\cellcolor{tab_red}24.52 &\cellcolor{tab_red}26.74 &\cellcolor{tab_red}0.331 &\cellcolor{tab_red}0.737 &\cellcolor{tab_red}0.729\\

\midrule
 
 \multirow{5}{*}{10k} & NeRF~\cite{mildenhall2021nerf} & \cellcolor{tab_orange}27.66 & \cellcolor{tab_orange}28.80 & \cellcolor{tab_yellow}0.425 & \cellcolor{tab_yellow}0.756 & \cellcolor{tab_yellow}0.749 & 23.63 & 24.90 & 0.458 & 0.686 & 0.650 & 22.78 & 24.49 & 0.538 & 0.663 & 0.638 \\
 & mip-NeRF 360~\cite{Barron_2022_CVPR} & \cellcolor{tab_yellow}27.41 & \cellcolor{tab_yellow}28.47 & \cellcolor{tab_orange}0.412 & \cellcolor{tab_orange}0.763 & \cellcolor{tab_orange}0.755 & \cellcolor{tab_orange}25.57 & \cellcolor{tab_orange}26.80 & \cellcolor{tab_orange}0.306 & \cellcolor{tab_orange}0.769 & \cellcolor{tab_orange}0.741 & \cellcolor{tab_orange}24.28 & \cellcolor{tab_orange}26.28 & \cellcolor{tab_orange}0.384 & \cellcolor{tab_orange}0.725 & \cellcolor{tab_orange}0.710  \\
 & TensoRF~\cite{chen2022tensorf} & 26.96 & 26.98 & 0.469 & 0.751 & 0.743 & \cellcolor{tab_yellow}24.09 & \cellcolor{tab_yellow}25.71 & \cellcolor{tab_yellow}0.436 & \cellcolor{tab_yellow}0.696 & \cellcolor{tab_yellow}0.676 & \cellcolor{tab_yellow}23.82 & \cellcolor{tab_yellow}25.75 & \cellcolor{tab_yellow}0.481 & \cellcolor{tab_yellow}0.694 & \cellcolor{tab_yellow}0.678  \\
 & DVGO~\cite{Sun_2022_CVPR} & 25.44 & 26.53 & 0.556 & 0.715 & 0.699 & 22.54 & 24.06 & 0.518 & 0.659 & 0.628 & 23.08 & 25.28 & 0.529 & 0.678 & 0.664\\
 & EgoNeRF & \cellcolor{tab_red}30.23 & \cellcolor{tab_red}31.47 & \cellcolor{tab_red}0.248 & \cellcolor{tab_red}0.840 & \cellcolor{tab_red}0.841 & \cellcolor{tab_red}28.81 & \cellcolor{tab_red}30.21 & \cellcolor{tab_red}0.136 & \cellcolor{tab_red}0.868 & \cellcolor{tab_red}0.859 & \cellcolor{tab_red}24.71 & \cellcolor{tab_red}26.98 & \cellcolor{tab_red}0.314 & \cellcolor{tab_red}0.746 & \cellcolor{tab_red}0.740 \\
 
 \midrule
 \multirow{5}{*}{100k} & NeRF~\cite{mildenhall2021nerf} & \cellcolor{tab_orange}31.67 & \cellcolor{tab_orange}33.08 & \cellcolor{tab_yellow}0.240 & \cellcolor{tab_yellow}0.852 & \cellcolor{tab_yellow}0.853 & \cellcolor{tab_yellow}27.12 & \cellcolor{tab_yellow}28.54 & \cellcolor{tab_yellow}0.269 & \cellcolor{tab_yellow}0.789 & \cellcolor{tab_yellow}0.772 & 24.91 & 26.65 & 0.384 & 0.721 & 0.702  \\
 & mip-NeRF 360~\cite{Barron_2022_CVPR} & \cellcolor{tab_yellow}31.12 & \cellcolor{tab_yellow}32.41 & \cellcolor{tab_orange}0.225 & \cellcolor{tab_orange}0.859 & \cellcolor{tab_orange}0.859 & \cellcolor{tab_orange}29.34 & \cellcolor{tab_orange}30.63 & \cellcolor{tab_orange}0.135 & \cellcolor{tab_orange}0.879 & \cellcolor{tab_orange}0.867 & \cellcolor{tab_red}25.57 & \cellcolor{tab_red}27.62 & \cellcolor{tab_red}0.268 & \cellcolor{tab_red}0.778 & \cellcolor{tab_red}0.770 \\
 & TensoRF~\cite{chen2022tensorf} & 29.25 & 30.57 & 0.376 & 0.791 & 0.793 & 25.68 & 27.47 & 0.344 & 0.734 & 0.726 & \cellcolor{tab_yellow}25.16 & 27.13 & \cellcolor{tab_yellow}0.376 & \cellcolor{tab_yellow}0.732 & 0.724 \\
 & DVGO~\cite{Sun_2022_CVPR} & 28.84 & 30.23 & 0.348 & 0.798 & 0.803 & 24.87 & 26.73 & 0.363 & 0.720 & 0.711 & 24.90 & \cellcolor{tab_yellow}27.28 & \cellcolor{tab_yellow}0.376 & \cellcolor{tab_yellow}0.732 & \cellcolor{tab_yellow}0.729 \\
 & EgoNeRF & \cellcolor{tab_red}33.11 & \cellcolor{tab_red}34.53 & \cellcolor{tab_red}0.142 & \cellcolor{tab_red}0.902 & \cellcolor{tab_red}0.904 & \cellcolor{tab_red}30.56 & \cellcolor{tab_red}32.04 & \cellcolor{tab_red}0.087 & \cellcolor{tab_red}0.904 & \cellcolor{tab_red}0.901 & \cellcolor{tab_orange}25.25 & \cellcolor{tab_orange}27.50 & \cellcolor{tab_orange}0.286 & \cellcolor{tab_orange}0.763 & \cellcolor{tab_orange}0.758 \\
\bottomrule
\end{tabular}
}
\caption{Quantitative results in outward-looking \textit{OmniBlender} and \textit{Ricoh360} dataset. Top results are colored as \colorbox{tab_red}{top1}, \colorbox{tab_orange}{top2}, and \colorbox{tab_yellow}{top3}.}
\label{tab:quantitative}
\end{table*}

\section{Experiments}
\label{sec:experiments}

We demonstrate that EgoNeRF can quickly capture and synthesize novel views of large-scale scenes. 
We describe full implementation details including hyperparameter setup in the supplementary material.

\paragraph{Datasets}
Since many of the existing datasets for NeRF are dedicated to a setup where a bounded object is captured from outside-in viewpoints, we propose new synthetic and real datasets of large-scale environments captured with omnidirectional videos.
\textit{OmniBlender} is a realistic synthetic dataset of 11 large-scale scenes with detailed textures and sophisticated geometries in both indoor and outdoor environments, 25 images for both train and test, respectively.
It consists of omnidirectional images along a relatively small circular camera path.
The spherical images are rendered using Blender's Cycles path tracing renderer~\cite{blender} with 2000$\times$1000 resolution.
\textit{Ricoh360} is a real-world 360$^\circ$ video dataset captured with a Ricoh Theta V camera with 1920$\times$960 resolution.
We record video on the circular path by rotating an omnidirectional camera fixed with a selfie stick as shown in \cref{fig:teaser} (a).
The dataset consists of 11 diverse indoor and outdoor scenes, 50 images for train and test, respectively.
With the benefit of the simple procedure, the whole data acquisition is finished in less than 5 seconds, which enables capturing the surrounding scene while it remains nearly static.
We obtain camera poses using SfM library OpenMVG~\cite{moulon2016openmvg}.
A detailed description of our dataset can be found in the supplementary material.

\paragraph{Baselines}
EgoNeRF is a variant of NeRF~\cite{mildenhall2021nerf}, which synthesizes novel views of the scene using the neural volume trained with multi-view images.
However, the original NeRF utilizes an MLP to represent the scene volume.
There also exists a recent variant called mip-NeRF 360~\cite{Barron_2022_CVPR}, which combines many techniques to increase the quality of the results, including the adaptation to unbounded scenes by warping space farther than a certain radius.
EgoNeRF employs feature grids and vector-matrix factorization in the balanced spherical grid. DVGO~\cite{Sun_2022_CVPR} exploits feature grids in a Cartesian coordinate with great acceleration, whereas TensoRF~\cite{chen2022tensorf} deploys factorization, also in Cartesian coordinate.
For all the methods, we train with the same ray batch size and the same number of feature grids (for DVGO and TensoRF) with one RTX-3090 GPU for a fair comparison.

\subsection{Quantitative Results}
The quantitative results are reported in mean PSNR, SSIM~\cite{wang2004image}, and LPIPS~\cite{zhang2018unreasonable} across test images in \textit{OmniBlender} and \textit{Ricoh360} dataset in \cref{tab:quantitative}.
Since equirectangular images in our datasets have distortion near poles, we additionally measure weighted-to-spherically uniform PSNR and SSIM~\cite{sun2017weighted} ($\text{PSNR}^{\text{WS}}$ and $\text{SSIM}^{\text{WS}}$), which place smaller weights near the poles when evaluating the metrics.

\Cref{tab:quantitative} shows that EgoNeRF outperforms all compared methods across all error metrics in the \textit{OmniBlender} dataset. 
With the efficient grid structure of EgoNeRF, the difference is more significant in earlier iterations.
Considering the time for each iteration, the efficiency gap is even more significant, which is also visualized in ~\cref{fig:time-PSNR}.
Our approach shows high performance even at the early 5k steps, which takes 10 minutes of wall-clock time. In contrast, mip-NeRF 360 needs approximately 8 hours to outperform our results at 5k steps.
In \textit{Ricoh360}, EgoNeRF surpasses other methods in 5k and 10k training steps, and shows comparable results in 100k steps.
However, our approach sometimes produces spotty artifacts in real-world datasets because the camera pose estimates can be erroneous.
On the other hand, MLP-based methods show blurry rendering, which sporadically achieves better scores in error metrics.
Such a phenomenon is prominent when the error in the camera pose makes the rays hit neighboring cells in the feature grid, which is further discussed in the supplementary material.

More importantly, the feature grid using the Cartesian coordinate system (TensoRF and DVGO) results in inferior performance, especially in outdoor scenes.
This supports our main claim that the Cartesian grid is inadequate to represent large-scale scenes captured from egocentric viewpoints.
In contrast, MLP-based methods (NeRF and mip-NeRF 360) achieve moderate results.

\begin{figure*}
    \centering
    \begin{tabular}{@{}c@{\,}c@{\,}c@{\,}|@{\,}c@{\,}c@{\,}}
    & \small \textit{BarberShop} & \small \textit{BistroBike} & \small \textit{Bricks} & \small \textit{Poster} \\
    
    \rotatebox{90}{\qquad\quad\:\;\: \small G.T.} & \includegraphics[width=0.23\linewidth]{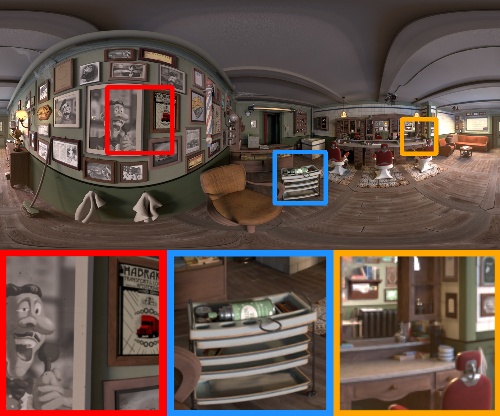} & \includegraphics[width=0.23\linewidth]{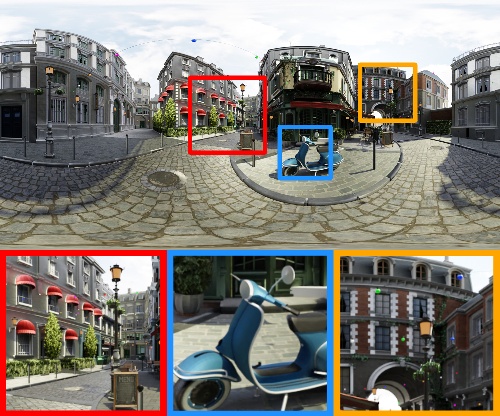}& \includegraphics[width=0.23\linewidth]{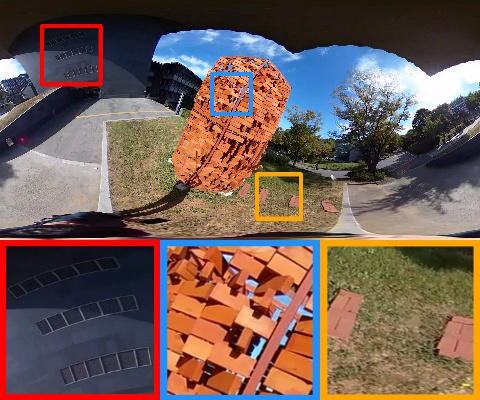}&\includegraphics[width=0.23\linewidth]{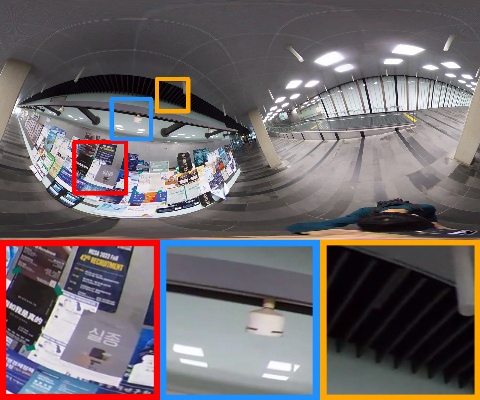}\\
    
    \rotatebox{90}{\qquad\;\, \small NeRF~\cite{mildenhall2021nerf}} & \includegraphics[width=0.23\linewidth]{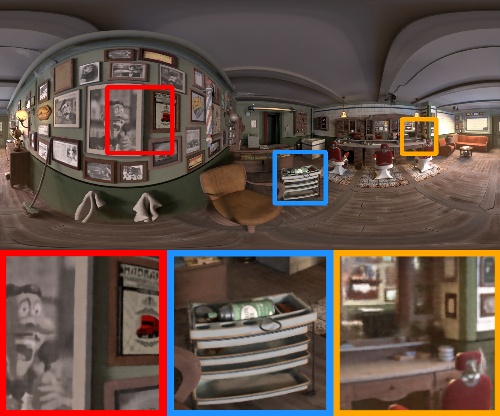} & \includegraphics[width=0.23\linewidth]{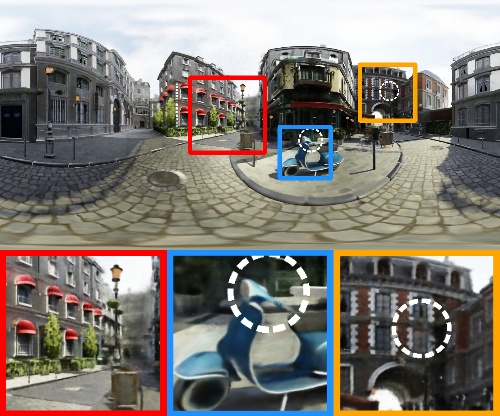}& \includegraphics[width=0.23\linewidth]{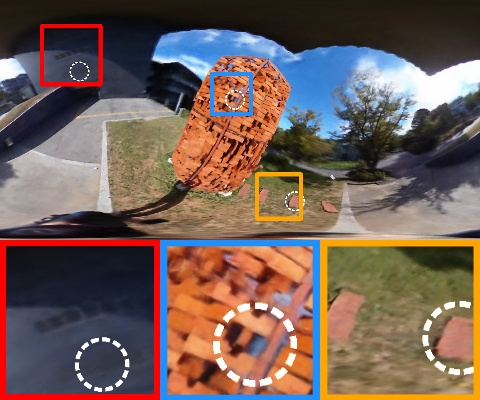}&\includegraphics[width=0.23\linewidth]{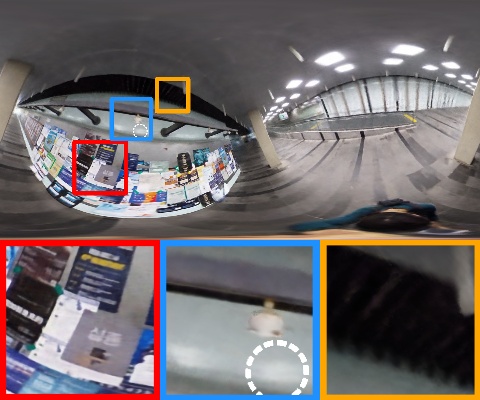}\\
    
    \rotatebox{90}{\quad\; \small mip-NeRF 360~\cite{Barron_2022_CVPR}} & \includegraphics[width=0.23\linewidth]{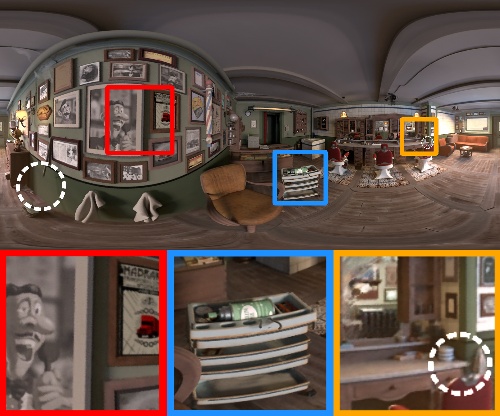} & \includegraphics[width=0.23\linewidth]{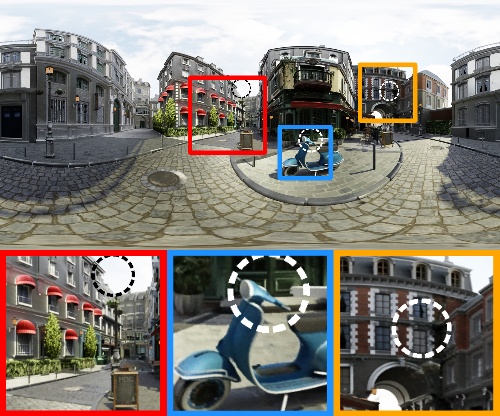}& \includegraphics[width=0.23\linewidth]{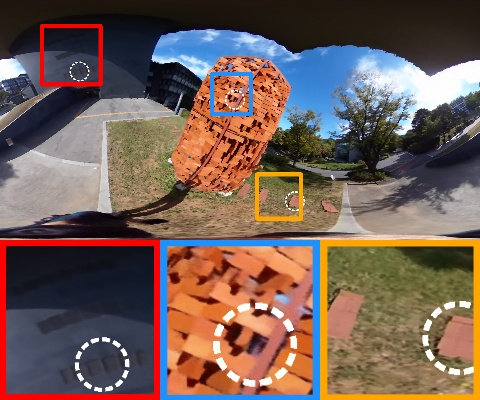}&\includegraphics[width=0.23\linewidth]{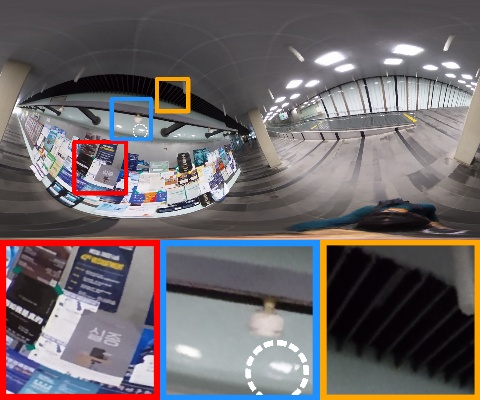}\\
    
    \rotatebox{90}{\qquad\; \small TensoRF~\cite{chen2022tensorf}} & \includegraphics[width=0.23\linewidth]{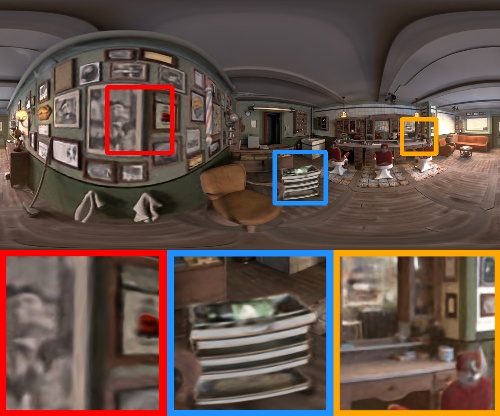} & \includegraphics[width=0.23\linewidth]{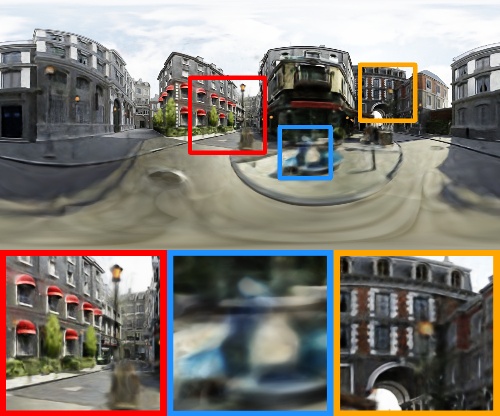}& \includegraphics[width=0.23\linewidth]{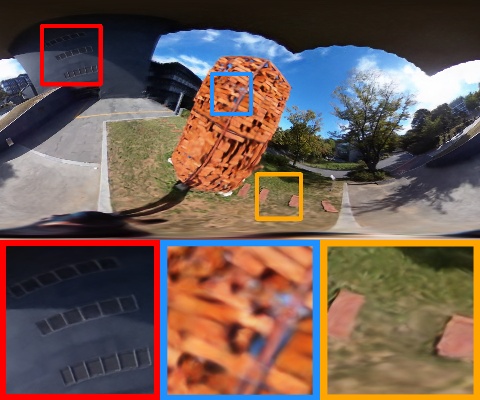}&\includegraphics[width=0.23\linewidth]{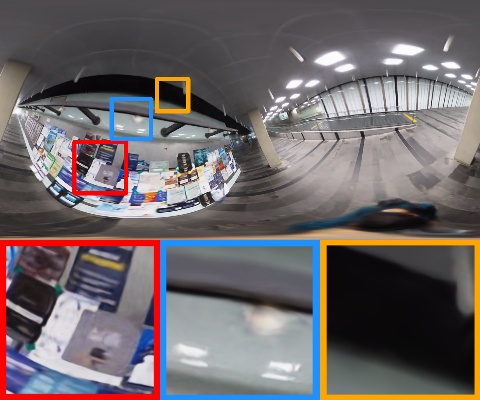}\\
    
    \rotatebox{90}{\qquad\:\: \small DVGO~\cite{Sun_2022_CVPR}} & \includegraphics[width=0.23\linewidth]{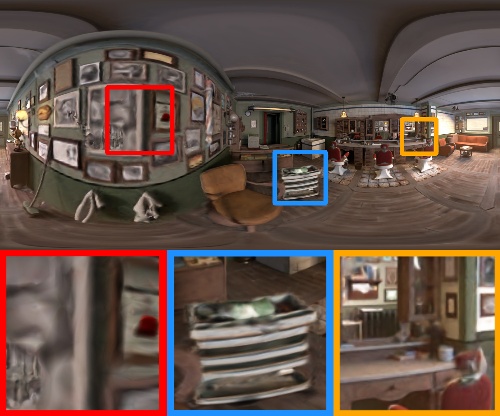} & \includegraphics[width=0.23\linewidth]{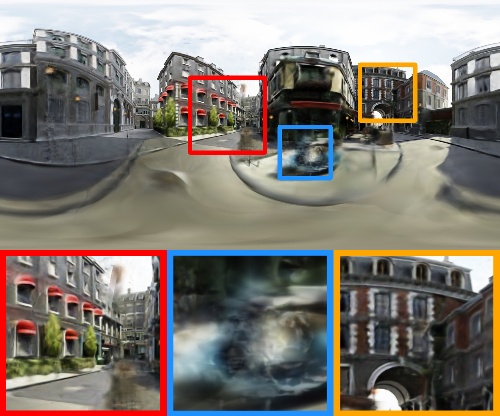}& \includegraphics[width=0.23\linewidth]{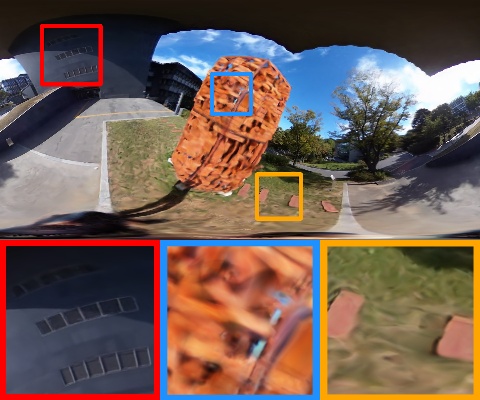}&\includegraphics[width=0.23\linewidth]{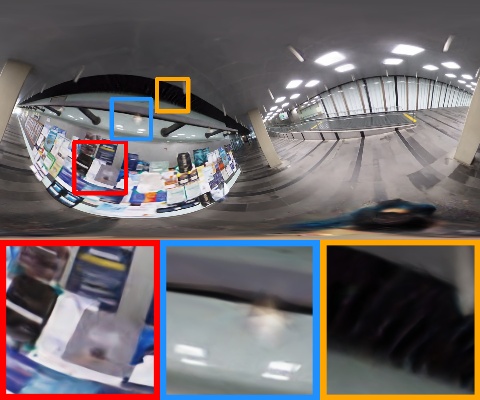}\\
    
    \rotatebox{90}{\qquad\quad \small EgoNeRF} & \includegraphics[width=0.23\linewidth]{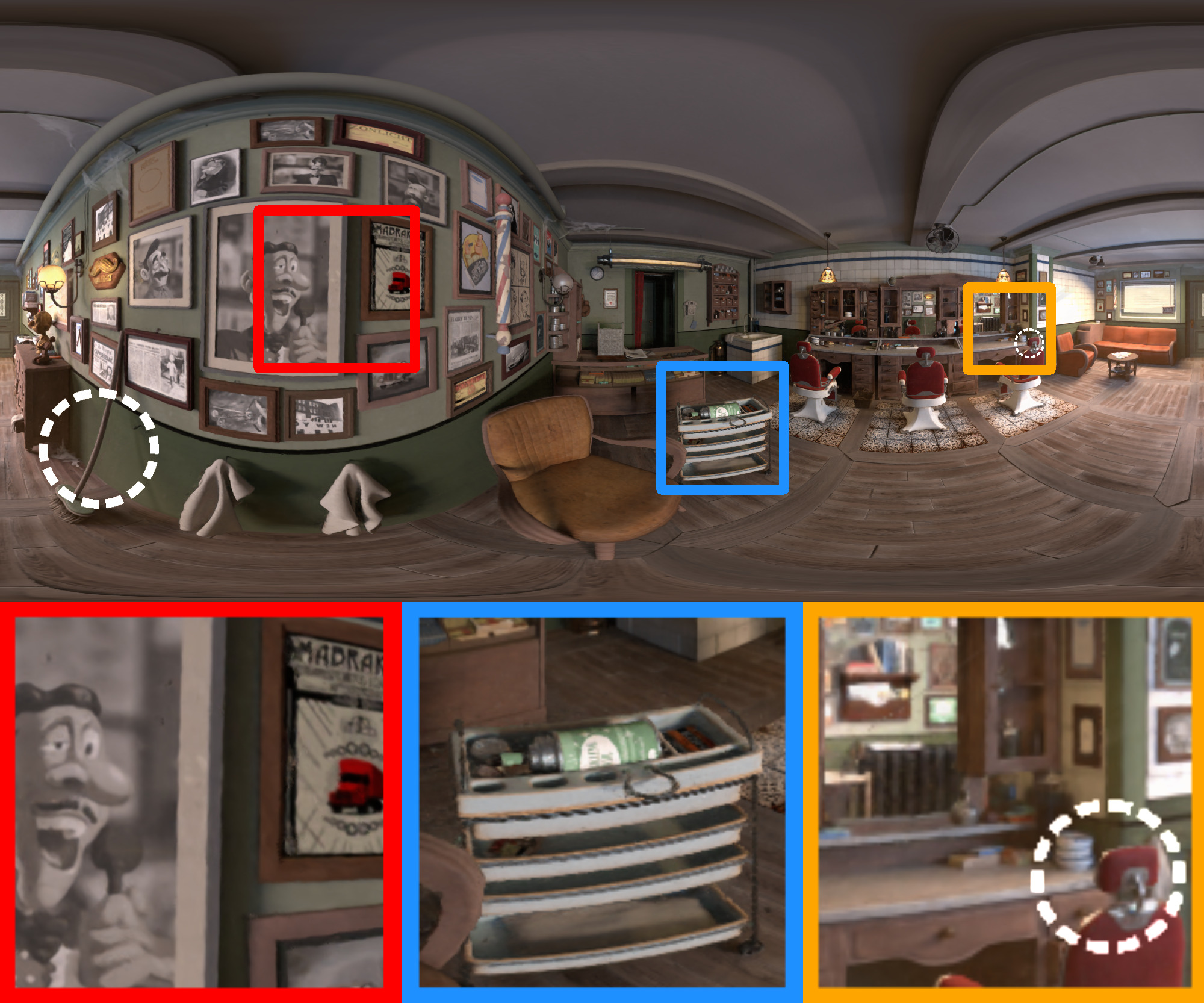} & \includegraphics[width=0.23\linewidth]{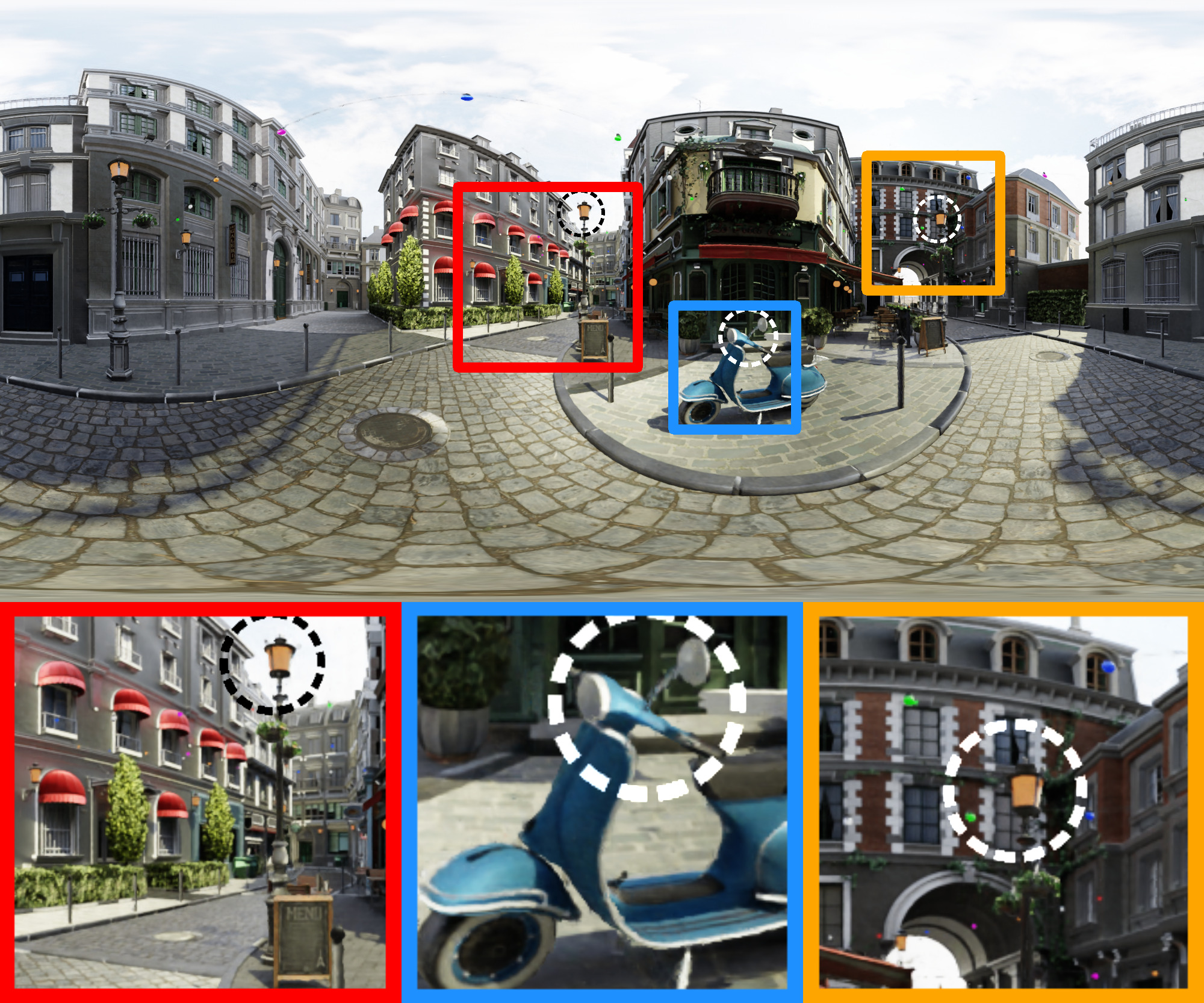}& \includegraphics[width=0.23\linewidth]{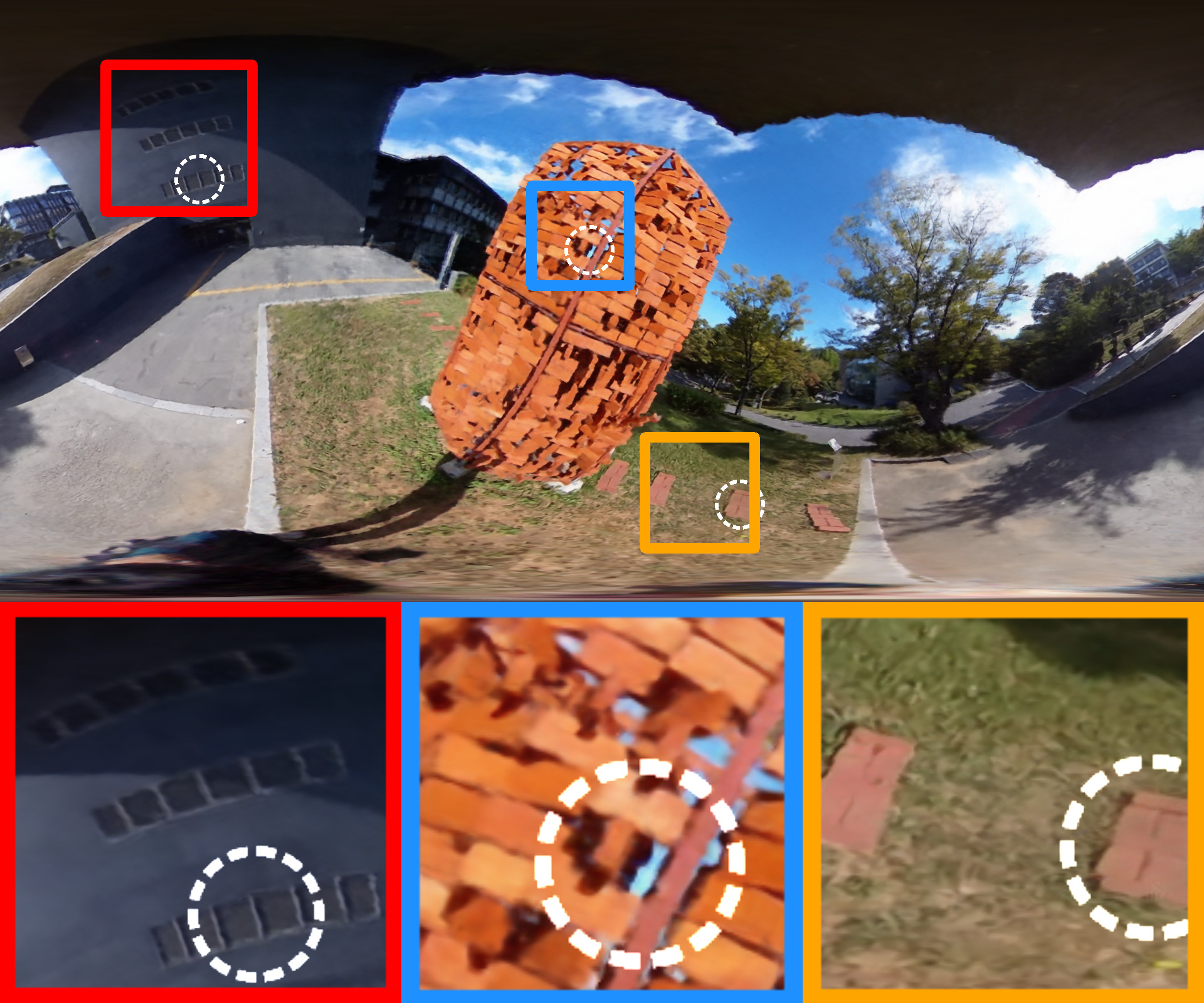}&\includegraphics[width=0.23\linewidth]{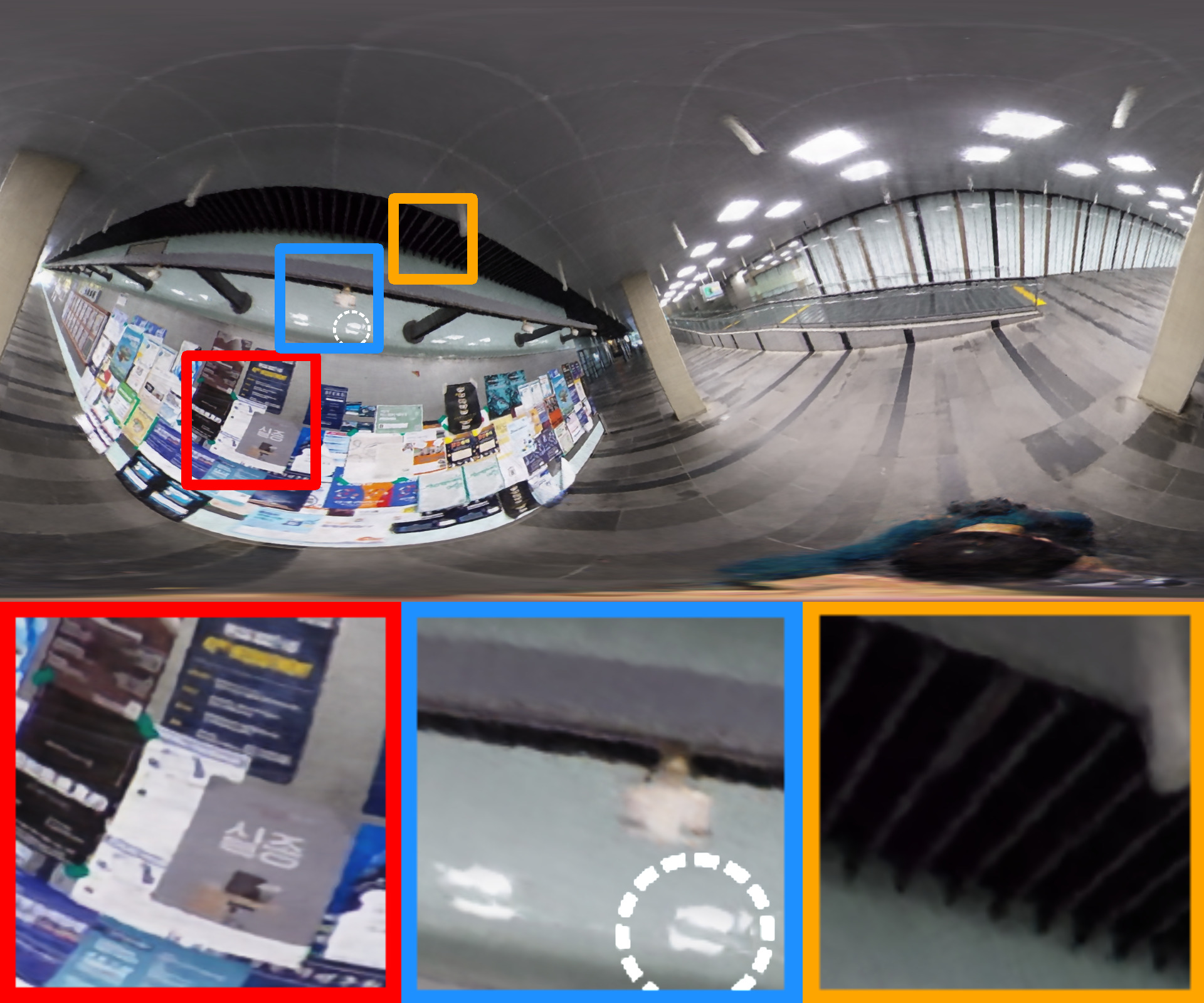}\\
    
    & \multicolumn{2}{c}{\small (a) \textit{OmniBlender}} & \multicolumn{2}{c}{\small (b) \textit{Ricoh360}} 
    
    \end{tabular}
    \vspace{-0.1in}
        
    \caption{Comparative results of novel view synthesis on the outward-looking (a) synthetic \textit{OmniBlender} dataset and (b) real-world \textit{Ricoh360} dataset. Best viewed on screen.}
    \label{fig:qualitative_comparison}
\end{figure*}

\subsection{Qualitative Results}

The qualitative results in \textit{OmniBlender} and \textit{Ricoh360} datasets are demonstrated in~\cref{fig:qualitative_comparison}.
Our method reconstructs fine details in both close-by objects and far-away regions.
However, for Cartesian grid-based methods (TensoRF and DVGO), many cells are wasted without being trained in far objects, while center cells might not have a sufficient resolution as depicted in \cref{fig:grid_comparison}.
It results in visible artifacts in the picture in \textit{BarberShop}, bike in \textit{BistroBike}, bricks in \textit{bricks}, and posters in \textit{poster}.
This phenomenon is predominant in large scenes, while EgoNeRF gives consistently faithful results regardless of the size of the scenes.

MLP-based approaches show better visual results than Cartesian feature grid-based methods with much longer training and rendering time.
However, mip-NeRF 360 often misses fine structures: e.g., stick of broom, thin handle and support fixture in tray, headrest attachment in chair in \textit{Barbershop}, street lamp, side mirror of bike, small colorful lightbulbs and thin wire in \textit{BistroBike}.
Some of them are also indicated with white dotted circles in \cref{fig:qualitative_comparison}. 
This may be due to mip-NeRF 360 resample ray samples from the proposal MLP and do not apply rendering loss for the proposal MLP to relieve lengthy training time to optimize large MLPs, thus the weight distribution is strongly determined by the initial guess of the proposal MLP.
In contrast, since our approach shares the same density grid $\mathcal{G}_\sigma$ to query volume density at coarse samples and fine samples, EgoNeRF shows superior rendering results on fine details.
Also, MLP-based approaches show smoothed rendering results across all the scenes (e.g., windows are blurred, cannot see the sky through the gap between bricks, the boundaries between stepping blocks are blurred in \textit{Bricks}, two reflected lights are merged in \textit{poster}. Some of them are also highlighted with white dotted circles in \cref{fig:qualitative_comparison}.), while EgoNeRF shows high-quality images similar to ground-truth images.

\subsection{Ablation study}
\label{subsec:ablation}

\begin{table}
    \centering
    \resizebox{0.9\linewidth}{!}{
    \begin{tabular}{@{}l|ccc|ccc@{}}
    \toprule
    \multirow{2}{*}{Method} & \multicolumn{3}{c|}{Indoor} & \multicolumn{3}{c}{Outdoor}\\
    & PSNR  & LPIPS & SSIM & PSNR & LPIPS & SSIM \\
    \midrule
    w/o exp $R$ grid & 31.32 & 0.188 & 0.871 & 26.66 & 0.187 & 0.792 \\
    w/o Yin-Yang grid & 30.53 & 0.191 & 0.860 & 26.74 & 0.160 & 0.806 \\
    Spherical Grid & 30.78 & 0.209 & 0.858 & 26.25 & 0.213 & 0.773 \\
    w/o Resampling & 32.40 & 0.167 & 0.886 & 30.12 & 0.105 & 0.891 \\
    w/o Environment map &  & - &  & 30.04 & 0.107 & 0.891 \\
    \midrule
    EgoNeRF (full) & 33.11 & 0.142 & 0.902 & 30.56 & 0.087 & 0.904\\
    \bottomrule
    \end{tabular}
    }
    \caption{An ablation study in \textit{OmniBlender} dataset. We replace and remove important components in EgoNeRF.}
    \label{tab:ablation}
\end{table}
\begin{figure}
    \centering
    \includegraphics[width=0.97\linewidth]{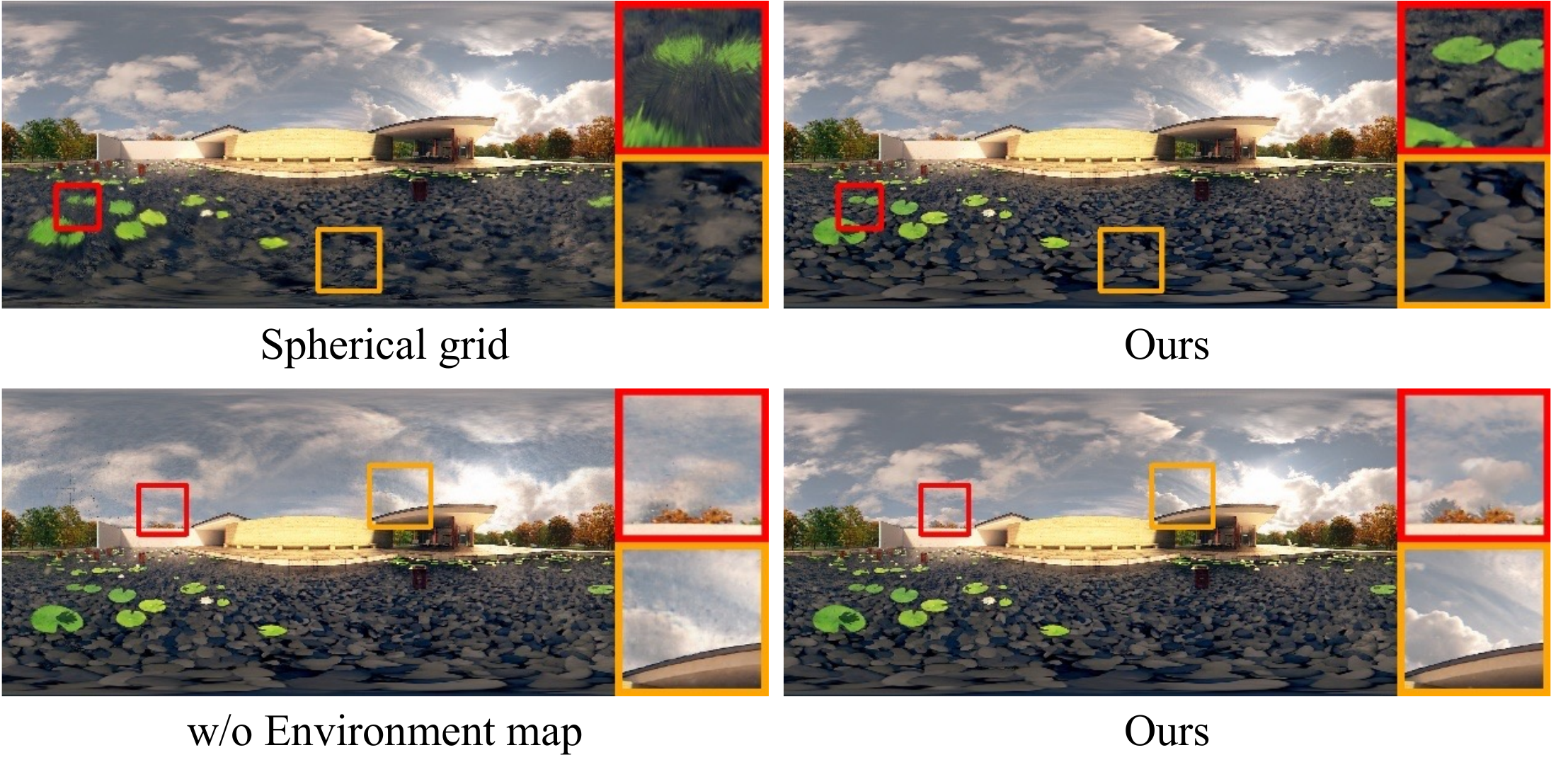}
    \caption{Qualitative results of ablation study.}
    \label{fig:ablation}
\end{figure}
We analyze the effects of important components of EgoNeRF with ablated versions.
\Cref{tab:ablation} shows that removing any of the components in our model degrades the performance across all metrics.
The first three rows are related to the balanced spherical grid.
Using the uniform radial partition deteriorates the performance, especially in outdoor scenes.
Without Yin-Yang grids, the angular partition exhibits high valence grid points on two poles and degrades the error metrics consequently. 
Removing both radial and angular balanced grids, which is identical to uniform spherical grids, causes the biggest drop in performance except PSNR in indoor scenes.
As shown in the first row of~\cref{fig:ablation}, the spherical feature grid has radial direction artifacts (red box) and shows blurrier rendered results for nearby objects compared to our full model.
Also, not employing resampling techniques and using a double number of ray samples reduces performance.
Lastly, removing the environment map in outdoor scenes shows blurry artifacts in infinitely far regions as shown in the second row of~\cref{fig:ablation} and reduces the performance consequently.

We provide additional analysis on the impact of hyperparameters, scene depths, and out-of-distribution testing in the supplementary material.

\section{Conclusion}

We present EgoNeRF, an efficient adaptation of the NeRF into large-scale scenes with casual input. 
We utilize a balanced spherical feature grid and maintain uniform ray hit rates for individual cells for scenes captured with a short video of omnidirectional cameras.
Together with factorization and resampling techniques, we can achieve fast and high-quality rendering of various indoor and outdoor environments.

Although EgoNeRF significantly outperforms the prior works in terms of visual quality and our approach converges much faster than MLP-based methods, we have some limitations.
In this paper, we do not consider all the challenges that come from real-world scenarios such as photometric variation from automatic camera exposure.
EgoNeRF sometimes shows noisy artifacts when the camera poses are not correct in the real-world \textit{Ricoh360} dataset, while MLP-based algorithms output blurred images.
Further analysis of the impact of camera parameter error is provided in the supplementary material.
One can resolve this by jointly optimizing the camera parameters as in~\cite{Lin_2021_ICCV, wang2021nerf, Jeong_2021_ICCV}.
Furthermore, like other NeRF-based models, we assume that scenes are static.

\paragraph{Acknowledgements}
This work was supported by the National Research Foundation of Korea (NRF) grant funded by the Korea government (MSIT) (No. RS-2023-00208197) and Institute of Information \& communications Technology Planning \& Evaluation (IITP) grant funded by the Korea government (MSIT) [NO.2021-0-01343, Artificial Intelligence Graduate School Program (Seoul National University)].
Young Min Kim is the corresponding author.

{\small
\bibliographystyle{ieee_fullname}
\bibliography{EgoNeRF_ref}
}

\clearpage
\noindent\textbf{\Large Supplementary material}

\appendix

\section{Vector-Matrix Decomposition of Balanced Spherical Feature Grids}
In this section, we describe the vector-matrix factorization of our balanced spherical feature grids.
As mentioned in Sec. 3.1. of the main manuscript, we model the radiance fields as 3D/4D tensors, which map a 3D position vector to volume density $\sigma$ and appearance feature vector.
Inspired by \cite{chen2022tensorf}, we decompose the 3D/4D tensor into low-rank tensor components.

\paragraph{VM Decomposition}
VM decomposition or vector-matrix decomposition, proposed by~\cite{chen2022tensorf}, decomposes a 3D tensor $\mathbf{T}\in \mathbb{R}^{I\times J\times K}$ into multiple vectors and matrices:
\begin{equation}
    \mathbf{T} = \sum_{n=1}^{N_1}\mathbf{v}_n^1 \otimes \mathbf{M}_n^{2,3}+\sum_{n=1}^{N_2}\mathbf{v}_n^2\otimes\mathbf{M}_n^{3,1}+\sum_{n=1}^{N_3}\mathbf{v}_n^3\otimes\mathbf{M}_n^{1,2},
\label{eq:vm_decomposition}
\end{equation}
where $\otimes$ denotes outer product, $\mathbf{v}_n^1\in\mathbb{R}^I$, $\mathbf{v}_n^2\in\mathbb{R}^J$, $\mathbf{v}_n^3\in\mathbb{R}^K$, and $\mathbf{M}_n^{2,3}\in\mathbb{R}^{J\times K}$, $\mathbf{M}_n^{3,1}\in\mathbb{R}^{K\times I}$, $\mathbf{M}_n^{1,2}\in\mathbb{R}^{I\times J}$ are vector and matrix factors for three modes of $n$th component respectively.
In general, $N_1, N_2, N_3$ have different values, but we use the same number of components for each mode for simplicity. i.e. $N_1=N_2=N_3=N$.
Then, \cref{eq:vm_decomposition} can be expressed as
\begin{equation}
    \mathbf{T} = \sum_{n=1}^N\sum_{m\in\{1, 2, 3\}}\mathcal{A}_{n}^m,
\label{eq:vm_decomposition_simple}
\end{equation}
where $\mathcal{A}_n^1 = \mathbf{v}_n^1\otimes\mathbf{M}_n^{2,3}$, $\mathcal{A}_n^2 = \mathbf{v}_n^2\otimes\mathbf{M}_n^{3,1}$, $\mathcal{A}_n^3 = \mathbf{v}_n^3\otimes\mathbf{M}_n^{1,2}$.

\paragraph{VM Decomposition of Balanced Spherical Feature Grids}
Our density feature grid $\mathcal{G}_\sigma$ is a 3D tensor of $\mathbb{R}^{2N_r^y\times N_\theta^y \times N_\phi^y}$.
The overset grid $\mathcal{G}_\sigma$ is a union of two tensors $\mathcal{G}_\sigma^{\text{Yin}}$ and $\mathcal{G}_\sigma^{\text{Yang}}\in\mathbb{R}^{N_r^y\times N_\theta^y \times N_\phi^y}$.
Each 3D tensor is further decomposed into vector and matrix factors using \cref{eq:vm_decomposition_simple}:
\small
\begin{align}
    \mathcal{G}_\sigma^y &\seq \sum_{n=1}^{N_\sigma}\mathbf{v}_{\sigma,n}^{y,R}\sotimes\mathbf{M}_{\sigma,n}^{y,\Theta\Phi}\splus\mathbf{v}_{\sigma,n}^{y,\Theta}\sotimes\mathbf{M}_{\sigma,n}^{y,\Phi R}\splus\mathbf{v}_{\sigma,n}^{y,\Phi}\sotimes\mathbf{M}_{\sigma,n}^{y,R\Theta}\nonumber\\
    &\seq \sum_{n=1}^{N_\sigma} \sum_{m\in{R\Theta\Phi}} \mathcal{A}_{\sigma, n}^{y, m}\text{,}\quad y\in\{\text{Yin, Yang}\}\text{.}
\end{align}
\normalsize

In contrast, our appearance gird $\mathcal{G}_a\in\mathbb{R}^{2N_r^y\times N_\theta^y\times N_\phi^y\times C}$ is a 4D tensor which has additional $C$-dimensional neural appearance features.
Since the mode of appearance feature does not need high dimension as spatial modes ($R,\Theta,\Phi$), we assign only vector components $\mathbf{b}$ for this mode, instead of matrix components from~\cite{chen2022tensorf}.
Specifically, $\mathcal{G}_a$ also consists of two tensors $\mathcal{G}_a^{\text{Yin}}$ and $\mathcal{G}_a^{\text{Yang}}\in\mathbb{R}^{N_r^y\times N_\theta^y\times N_\phi^y \times C}$ and each are factorized as following:
\small
\begin{align}
    \mathcal{G}_a^y &\seq \sum_{n=1}^{N_a}\mathbf{v}_{a,n}^{y,R}\sotimes \mathbf{M}_{a,n}^{y,\Theta\Phi}\sotimes\mathbf{b}_{3n-2}^y \splus\mathbf{v}_{a,n}^{y,\Theta}\sotimes \mathbf{M}_{a,n}^{y,\Phi R}\sotimes\mathbf{b}_{3n-1}^y \nonumber\\
    &\qquad \splus\mathbf{v}_{a,n}^{y,\Phi}\sotimes \mathbf{M}_{a,n}^{y,R\Theta}\sotimes\mathbf{b}_{3n}^y \nonumber\\
    &\seq \sum_{n=1}^{N_a} \mathcal{A}_{a,n}^{y,R}\sotimes \mathbf{b}_{3n-2}^{y} \splus \mathcal{A}_{a,n}^{y,\Theta}\sotimes \mathbf{b}_{3n-1}^{y}\splus\mathcal{A}_{a,n}^{y,\Phi}\sotimes \mathbf{b}_{3n}^{y}.
    \label{eq:vm_decomposition_appearance}
\end{align}
\normalsize
$\mathbf{B}\in\mathbb{R}^{C\times6N_a}$ in Fig.4 of the main manuscript is a matrix obtained by stacking all $\mathbf{b}^y$s columnwise.
By using $\mathbf{B}$ matrix, we can calculate \cref{eq:vm_decomposition_appearance} with simple matrix multiplication.

\paragraph{Querying Values from Grids}
In the volume rendering pipeline, the volume density $\sigma$ and color $c$ are queried from our feature grids along the camera rays:
\begin{equation}
    \sigma(\mathbf{x})=\mathcal{T}(\mathcal{G}_\sigma,\mathbf{x}), \quad c(\mathbf{x}, \mathbf{d})=f_{\text{MLP}}(\mathcal{T}(\mathcal{G}_a, \mathbf{x}),\mathbf{d}),
\end{equation}
where $\mathbf{x}$, $\mathbf{d}$ are querying position and viewing direction respectively, and $\mathcal{T}$ is a trilinear interpolation operator, as denoted in Eq. (5) of the main manuscript.
Furthermore, we can reduce computational burden by replacing trilinear interpolation with linear/bilinear interpolation of vector/matrix factors.
\begin{figure*}
    \centering
    \includegraphics[width=\linewidth]{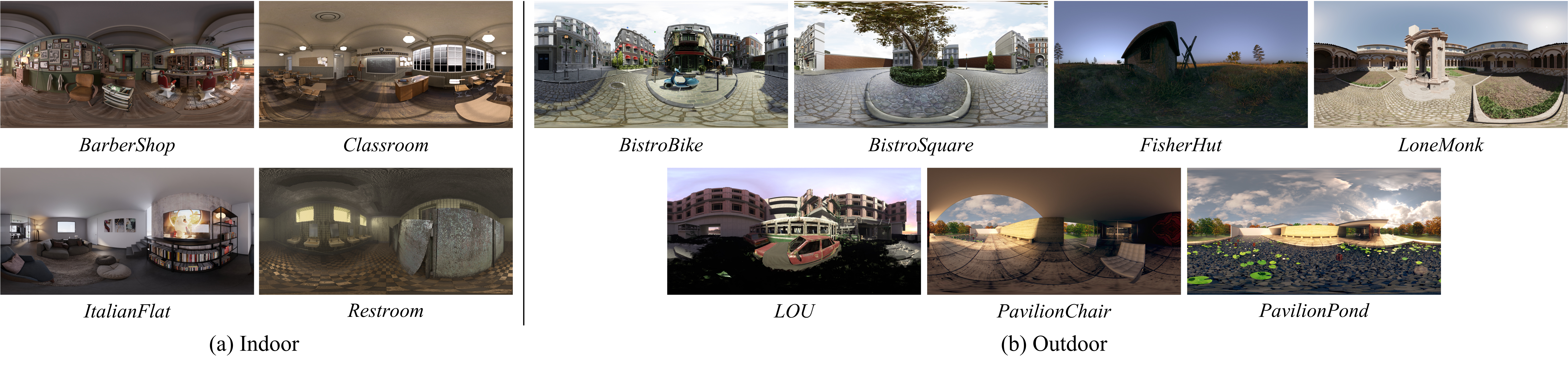}
    \caption{Samples from our synthetic \textit{OmniBlender} dataset.}
    \label{fig:dataset_omniblender}
\end{figure*}
\begin{figure*}
    \centering
    \includegraphics[width=\linewidth]{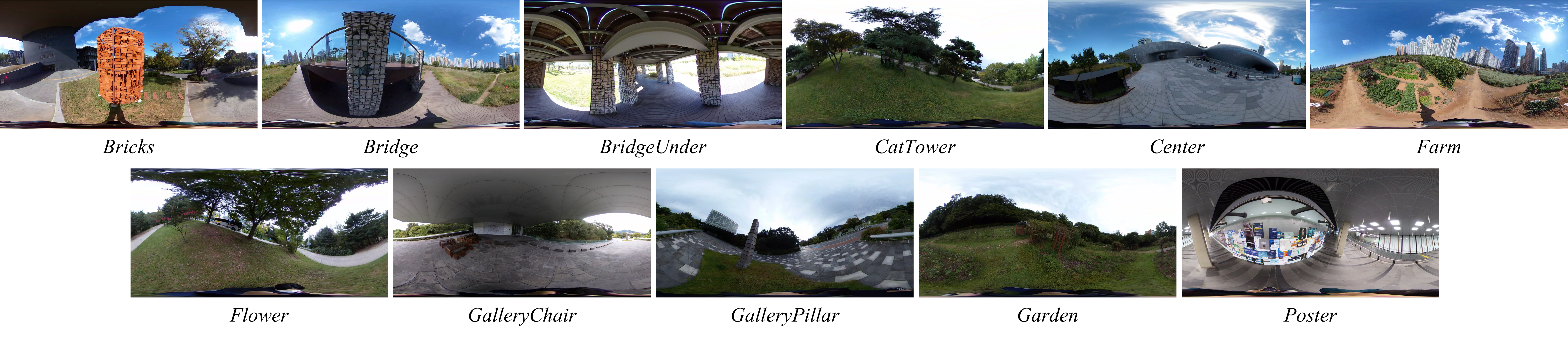}
    \caption{Samples from our real-world \textit{Ricoh360} dataset.}
    \label{fig:dataset_ricoh}
\end{figure*}

\section{Implementation Details}
\label{sec:implementastion}

EgoNeRF is implemented with PyTorch~\cite{paszke2019pytorch} without using any customized CUDA kernels. We will release the code and the dataset publicly upon publication.

\subsection{Hyperparameter Setup}
In this section, we report the hyperparameter setup used in experiments for EgoNeRF.
We use Adam optimizer~\cite{KingmaB14} with a learning rate of 0.02 following~\cite{chen2022tensorf}, and use default values of other hyperparameters of Adam optimizer ($\beta_1 = 0.9, \beta_2 = 0.999, \epsilon = 10^{-7}$).
For all scenes, we use $300^3$ voxels for both $\mathcal{G}_\sigma$ and $\mathcal{G}_a$ with 
$N_r^y:N_\theta^y:N_\phi^y=1:\frac{2\sqrt{3}}{3}:2\sqrt{3}$.
The dimension of appearance feature $C$ is 27 and we use two-layer MLP of 128 hidden units for decoding network $f_{\text{MLP}}$.
The number of decomposed components $N_\sigma=16$ and $N_a=48$.
We use the $r_0=0.03, 0.05$, and $R_{\text{max}}=15, 300$ for the indoor and outdoor scenes, respectively.
And the size of convolution kernel $K$ for obtaining a coarse grid is 2.

\subsection{Dataset Information}
\label{subsec:data}
\paragraph{OmniBlender}

\textit{OmniBlender} is our newly introduced synthetic dataset.
OmniBlender contains outward-looking images of 11 challenging large-scale indoor/outdoor environments with various objects and materials.
We render equirectangular images along short helix camera trajectory with Blender's Cycles path tracer~\cite{blender}.
All the images have 2000$\times$1000 resolution and we use 25 images for the training set and test set respectively.
Sample images from our dataset are presented in \cref{fig:dataset_omniblender}.
We slightly modify the publicly available 3D models from various sources below:

\definecolor{codegreen}{rgb}{0,0.6,0}
\definecolor{backcolour}{rgb}{0.96,0.96,0.96}

\lstdefinestyle{egonerf}{
backgroundcolor=\color{backcolour},   
    commentstyle=\color{codegreen}\itshape,
    basicstyle=\ttfamily\scriptsize,
    breakatwhitespace=false,         
    breaklines=true,                 
    captionpos=b,                    
    keepspaces=true,                 
    numbers=none,                    
    numbersep=5pt,                  
    showspaces=false,                
    showstringspaces=false,
    showtabs=false,                  
    tabsize=2,
    emph={BarberShop, Classroom, ItalianFlat, Restroom, Bistro, FisherHut, LoneMonk, LOU, Pavilion},
    emphstyle=\bfseries
}
\lstset{style=egonerf}
\begin{lstlisting}[language=python]
# Indoor
- BarberShop by Blender (CC-BY)
https://www.blender.org/download/demo-files/
- Classroom by Christophe Seux (CC-BY)
https://www.blender.org/download/demo-files/
- ItalianFlat by Flavio Della Tommasa (CC-BY)
https://www.blender.org/download/demo-files/
- Restroom by oldtimer (CC-BY)
https://blendswap.com/blend/14216
# Outdoor
- Bistro by Amazon Lumberyard (CC-BY)
http://developer.nvidia.com/orca/amazon-lumberyard-bistro
- FisherHut by DHCG (CC-BY)
https://www.blendswap.com/blend/30099
- LoneMonk by Carlo Bergonzin (CC-BY)
https://www.blender.org/download/demo-files/
- LOU by Andreas Stromberg
https://stromberg.gumroad.com/l/dGeBoS
- Pavilion by eMirage (CC-BY)
https://www.blender.org/download/demo-files/
\end{lstlisting}

\paragraph{Ricoh360}
\textit{Ricoh360} is our newly introduced real-world dataset.
The dataset contains short omnidirectional videos captured by rotating a commercial Ricoh Theta V camera attached to a selfie stick.
We collect 11 large-scale scenes from various indoor/outdoor environments.
After capturing the videos, we estimate camera parameters corresponding to each images by structure from motion library~\cite{moulon2016openmvg}.
We use 50 images for training set and test set respectively.
Sample images from Ricoh360 dataset are demonstrated in \cref{fig:dataset_ricoh}.
\newcommand{\ssim}{\,{\sim}\,}
\section{Analysis on Camera Parameter Error}
\begin{figure}
    \centering
    \includegraphics[width=\linewidth]{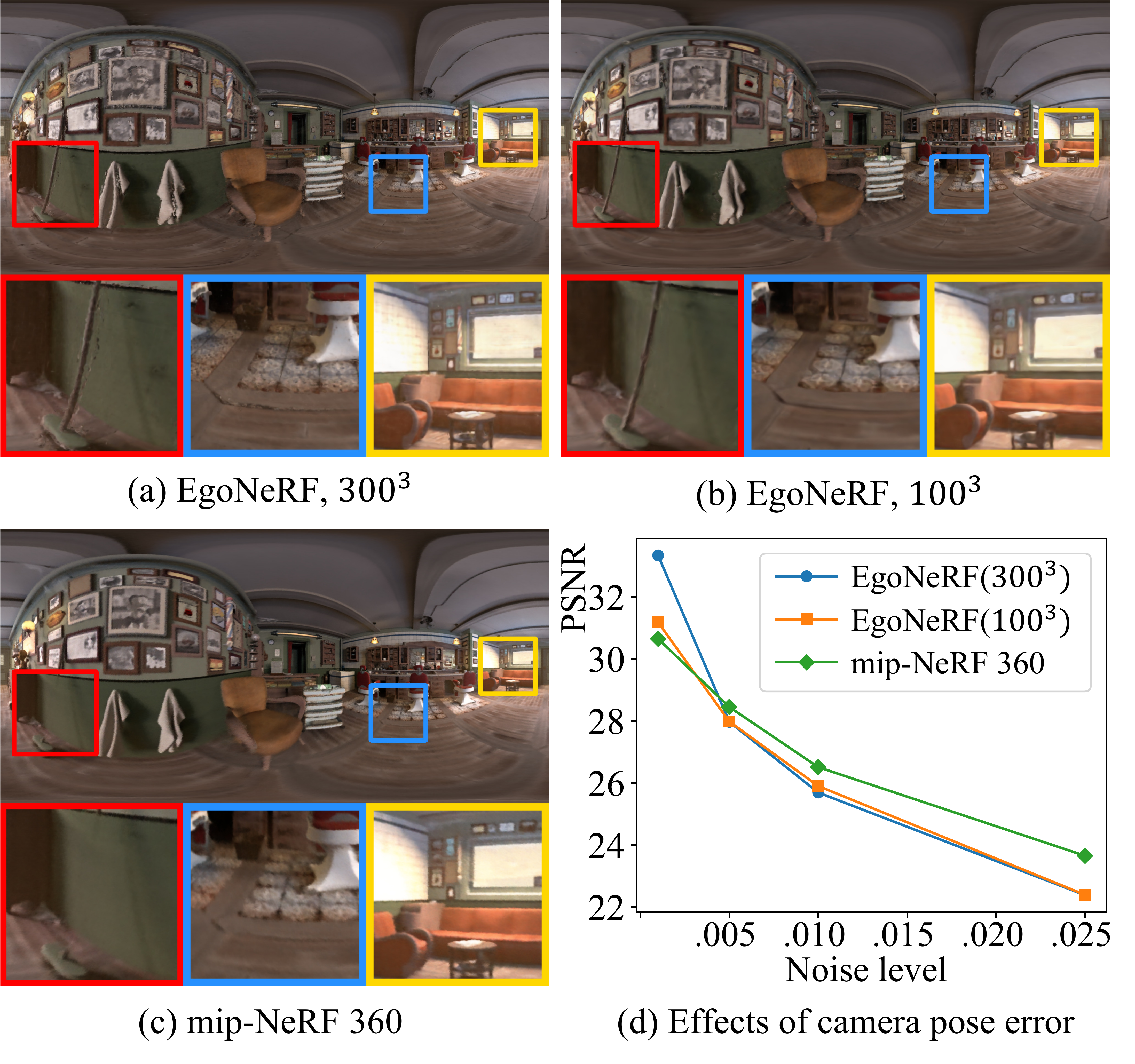}
    \caption{Qualitative results under the Gaussian perturbation $\epsilon\ssim(0,0.005)$ in different models (a) EgoNeRF, (b) EgoNeRF with coarser grid ($100^3$), and (c) MLP-based method mip-NeRF 360~\cite{Barron_2022_CVPR}. (d) Quantitative results of injecting different levels of Gaussian noise into camera poses.}
    \label{fig:noise_study}
\end{figure}
In this section, we analyze the effects of errors in camera parameters.
While EgoNeRF is able to reconstruct precise 3D scenes and synthesize high-quality novel views under perfect camera parameters in synthetic \textit{OmniBlender} dataset, our approach shows a degraded performance when the camera poses have errors in real-world scenes like prior works.
To further study the effects of the camera pose error, we train EgoNeRF with different grid resolutions ($300^3$ which is identical to our original setup, and $100^3$) and MLP-based approach mip-NeRF 360~\cite{Barron_2022_CVPR} under various levels of camera pose errors.
We perturb the camera pose by adding Gaussian noises $\epsilon\ssim(0,\sigma^2)$ with different levels of variance in the \textit{BarberShop} scene in \textit{OmniBlender}.

As shown in \cref{fig:noise_study} (d), injecting a higher level of noise reduces the performance across all the methods consistently.
EgoNeRF with the default parameter (resolution of $300^3$) outperforms other baselines amidst a negligible amount of noise (variance of 0.001), which is coherent with the main results.
When the level of noise increases, however, the performance of our model with fine resolution degrades rapidly and reaches a similar level of EgoNeRF with coarse resolution.
The MLP-based approach~\cite{Barron_2022_CVPR} shows better performance in the presence of high level of noise (greater than 0.01).

In \cref{fig:noise_study} (a) to (c), we visualize the qualitative results in the noise level 0.005 of which PSNR values from different methods are comparable.
As shown in the red box of \cref{fig:noise_study} (a), we observe a noisy artifact nearby the fine structure of the close objects in EgoNeRF.
On the other hand, EgoNeRF with coarse resolution does not show such a phenomenon in the red box of \cref{fig:noise_study} (b).
In contrast, EgoNeRF with both fine and coarse resolution does not make the noisy artifact in the far-away regions (yellow box).
We hypothesize that if the camera pose noise is non-negligible with relative to the grid size, the wrong camera parameters cause the camera ray for fine objects to hit the wrong neighborhood grids, which leads to multiple erroneous reconstructions of fine structures.
Since our distance-adaptive balanced spherical feature grid has a small grid size near the center and the grid has a larger volume at far regions, the noisy artifact only appears at the close region.
As shown in the blue box in \cref{fig:noise_study}, the MLP-based method shows blurry artifacts amidst the noise in the camera pose in contrast to the noisy artifact in grid-based methods.
This may be because MLP output naturally interpolates the values observed in the training set.
\vspace{-0.5em}
\section{Inward-facing Dataset}

\begin{table}
    \centering
    \resizebox{\linewidth}{!}{
    \begin{tabular}{lccc|ccc}
    \toprule
    \multicolumn{1}{l|}{Method} & PSNR & LPIPS & SSIM & PSNR & LPIPS & SSIM \\
    \midrule
    \multicolumn{1}{l|}{NeRF~\cite{mildenhall2021nerf}} & 31.01 & 0.081 & 0.947 & 24.85 & 0.426 & 0.659 \\
    \multicolumn{1}{l|}{mip-NeRF~\cite{Barron_2021_ICCV}} & 32.63 & 0.047 & 0.958 & \cellcolor{tab_yellow}25.12 & \cellcolor{tab_yellow}0.414 & \cellcolor{tab_yellow}0.672 \\
    \multicolumn{1}{l|}{mip-NeRF 360~\cite{Barron_2022_CVPR}} & \cellcolor{tab_red}33.25 & 0.039 & \cellcolor{tab_orange}0.962 & \cellcolor{tab_red}29.23 & \cellcolor{tab_red}0.207 & \cellcolor{tab_red}0.844 \\
    \multicolumn{1}{l|}{TensoRF~\cite{chen2022tensorf}} & 
    \cellcolor{tab_orange}33.14 & \cellcolor{tab_red}0.027 & 
    \cellcolor{tab_red}0.963 & 22.75 & 0.619 & 0.558 \\
    \multicolumn{1}{l|}{DVGO~\cite{Sun_2022_CVPR}} & \cellcolor{tab_yellow}32.80 & \cellcolor{tab_red}0.027 & \cellcolor{tab_yellow}0.961 & 20.67 & 0.490 & 0.575 \\
    \multicolumn{1}{l|}{EgoNeRF} & 31.51 & \cellcolor{tab_yellow}0.037 & 0.952 & \cellcolor{tab_orange}25.83 & \cellcolor{tab_orange}0.320 & \cellcolor{tab_orange}0.701 \\
    \bottomrule
     & \multicolumn{3}{c}{(a) Synthetic-NeRF} & \multicolumn{3}{c}{(b) mip-NeRF 360}
    \end{tabular}
    }
    \caption{Quantitative results in inward-facing datasets (a) Synthetic-NeRF~\cite{mildenhall2021nerf} and (b) mip-NeRF 360~\cite{Barron_2022_CVPR}.}
    \label{tab:inward_facing}
\end{table}
The spherical grid of EgoNeRF aligns nicely with outward-facing scenes, not inward-facing images of typical NeRF settings.
We optionally report results from widely-used datasets for novel view synthesis in ~\cref{tab:inward_facing}.
EgoNeRF shows comparable results in the Synthetic-NeRF dataset~\cite{mildenhall2021nerf}, which contains 8 synthetic objects.
In mip-NeRF 360 dataset~\cite{Barron_2022_CVPR}, which contains inward-facing objects but has unbounded background scenes, EgoNeRF outperforms other baselines except mip-NeRF 360.
\section{Robustness Study}

\begin{table}
    \centering
    \resizebox{\linewidth}{!}{
    \begin{tabular}{@{}l|cccccccc@{}}
    \toprule
    $r_0$ (m) & 0.01 & 0.03 & $0.05^*$ & 0.1 & 0.15 & 0.2 & 0.3 & 0.5 \\
    PSNR & 32.95 & 33.70 & 34.07 & 34.50 & 34.39 & 33.93 & 33.34 & 31.80\\
    \midrule
    $R_{\text{max}}$ (m) & 25 & 50 & 100 & 200 & $300^{*}$ & 500 & 1000 & \\
    PSNR & 28.43 & 33.96 & 34.31 & 34.23 & 34.07 & 33.93 & 33.69 & \\
    \bottomrule
    \end{tabular}
    }
    \caption{Quantitative results in \textit{BistroBike} (max depth$\seq 220$) for different hyperparameters. We mark * for the default values.}
    \label{tab:hyperparameter}
\end{table}

\begin{table}
    \centering
    \resizebox{\linewidth}{!}{
    \begin{tabular}{@{}l|ccccccccc@{}}
    \toprule
    $r$ (m) & 0 & 0.1 & 0.2 & 0.5 & $1^*$ & 1.5 & 2 & 3 & 5 \\
    PSNR & 35.38 & 35.47 & 35.45 & 35.26 & 34.74 & 33.15 & 31.17 & 27.36 & 22.02\\
    \bottomrule
    \end{tabular}
    }
    \caption{Out of distribution test. $r$ is the distance between the center of training camera trajectory and position of test view.}
    \label{tab:ood}
\end{table}
In this section, we analyze the effects of various components to the reconstruction quality.
In ~\cref{tab:hyperparameter}, we study the effects of hyperparameters.
EgoNeRF shows robust performance regardless of the choice of $r_0$ and $R_{\text{max}}$ unless $R_{\text{max}}$ is too small compared to the scene size.

\begin{figure}
    \centering
    \includegraphics[width=\linewidth]{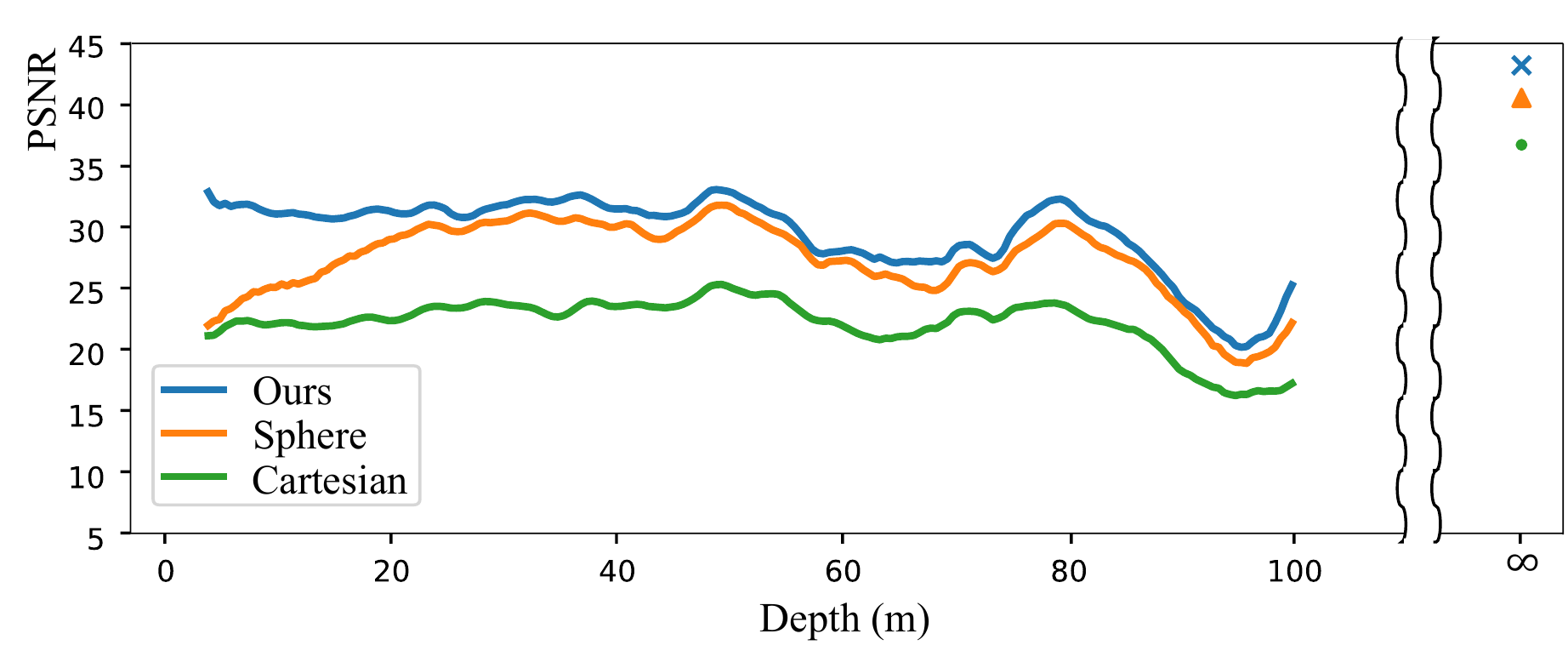}
    \caption{Reconstruction quality according to scene depth.}
    \label{fig:depth_study}
\end{figure}

Also, we compare the quality of reconstructed images at various depths in ~\cref{fig:depth_study}.
EgoNeRF outperforms regular spherical grid and Cartesian grid, especially in the near region.
It supports our claim that Cartesian grid or regular spherical grid has insufficient resolution at nearby regions and is extravagant for far objects.

We further provide the quality of rendering at various distances from the original trajectory ($r=1$) in ~\cref{tab:ood}.
We noticed only minimal quality degradation for $r<1$ and $1<r\leq 3$.
Only when the viewing position is extremely far ($r\geq5$), there exists a noticeable performance decrease due to the unseen regions and the unconstrained scene depth.
\begin{figure}
    \centering
    \includegraphics[width=\linewidth]{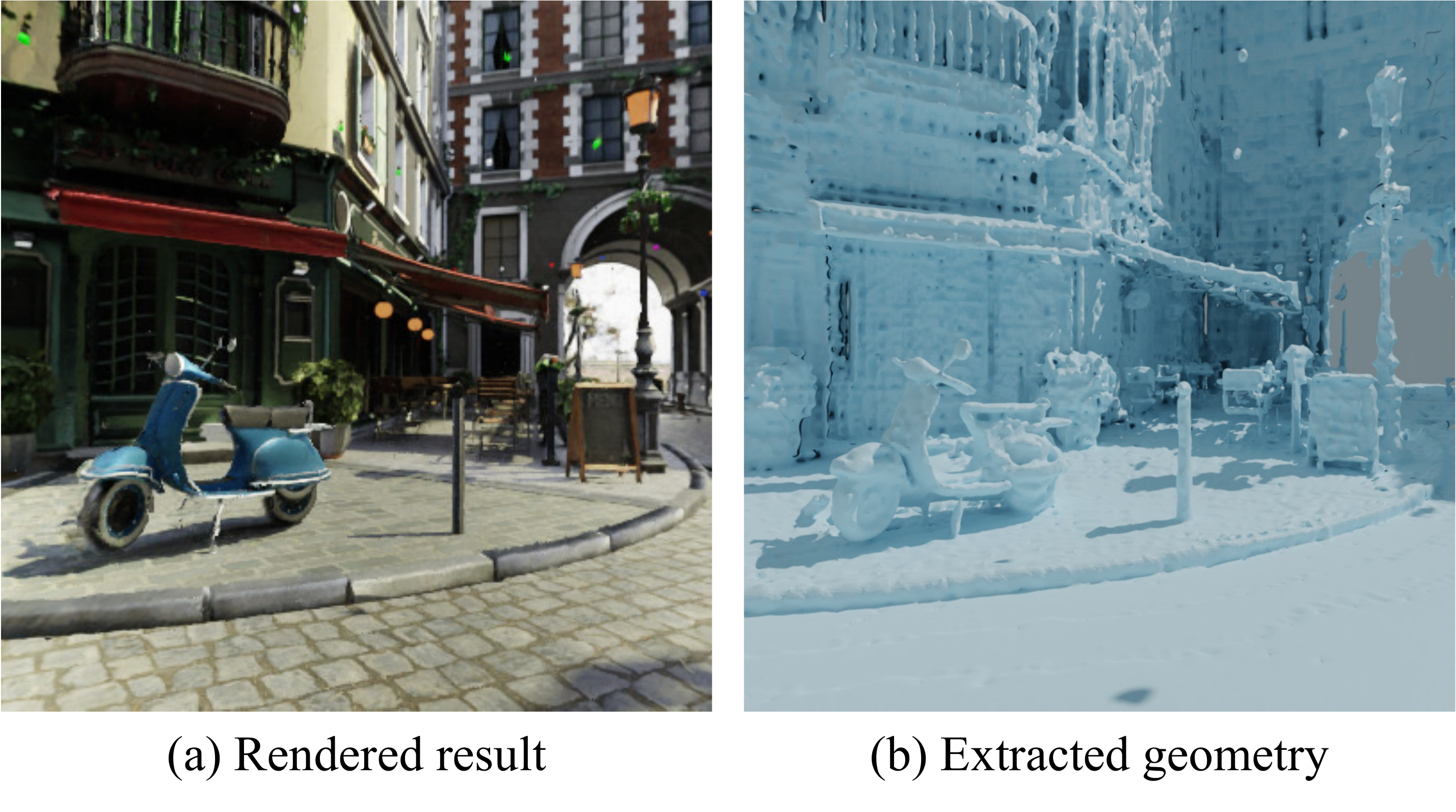}
    \caption{We demonstrate the (a) rendered result from novel viewpoint and (b) the extracted geometry from our density grid $\mathcal{G}_\sigma$ in \textit{BistroBike} scene in \textit{OmniBlender} dataset.}
    \label{fig:bistro_bike_geometry}
\end{figure}

\section{Additional Results}
\begin{figure*}
\centering
\includegraphics[width=\linewidth]{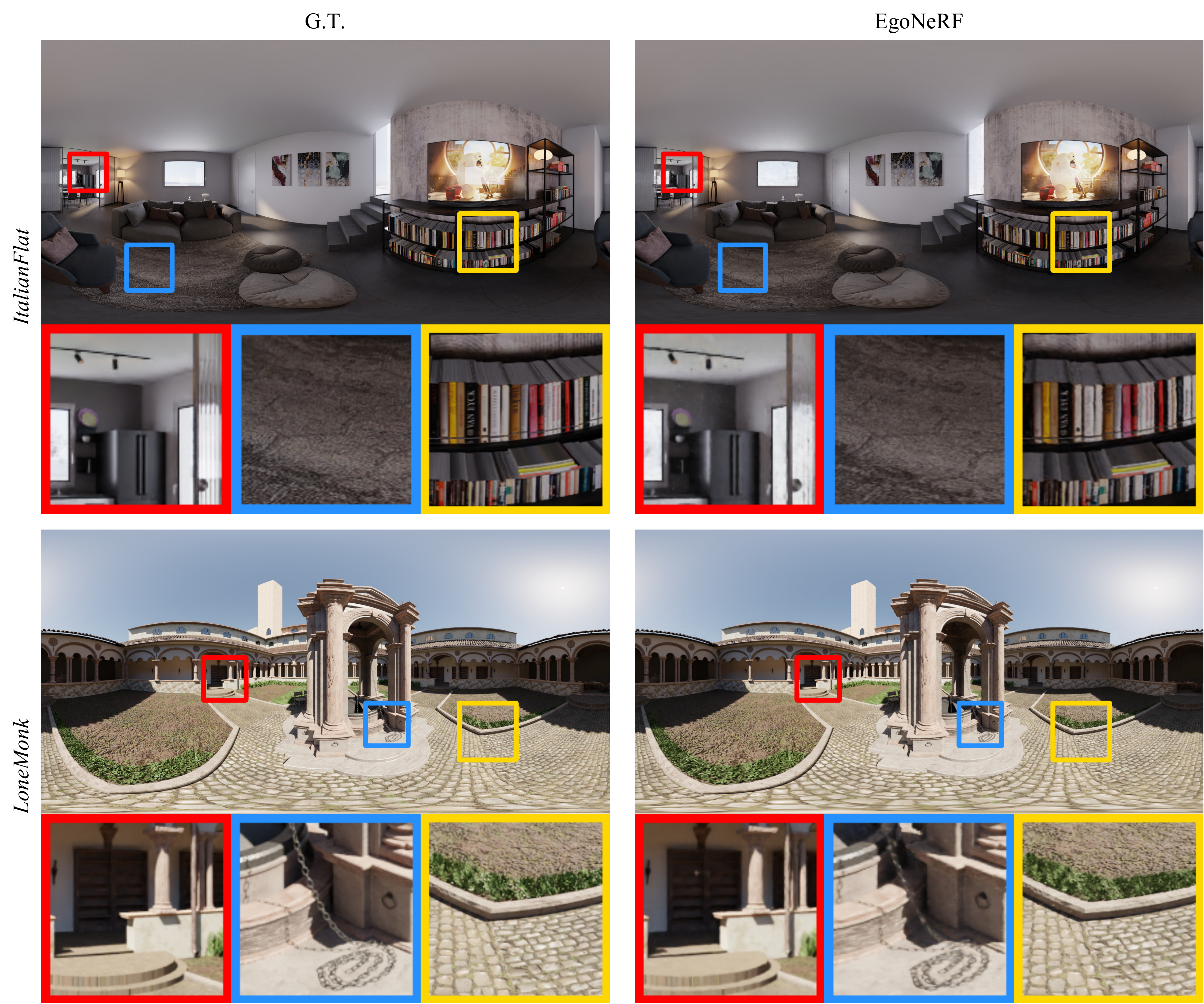}
\caption{Additional qualitative results in \textit{OmniBlender} dataset.}
\label{fig:additional_omniblender}
\end{figure*}

\begin{figure*}
\centering
\includegraphics[width=\linewidth]{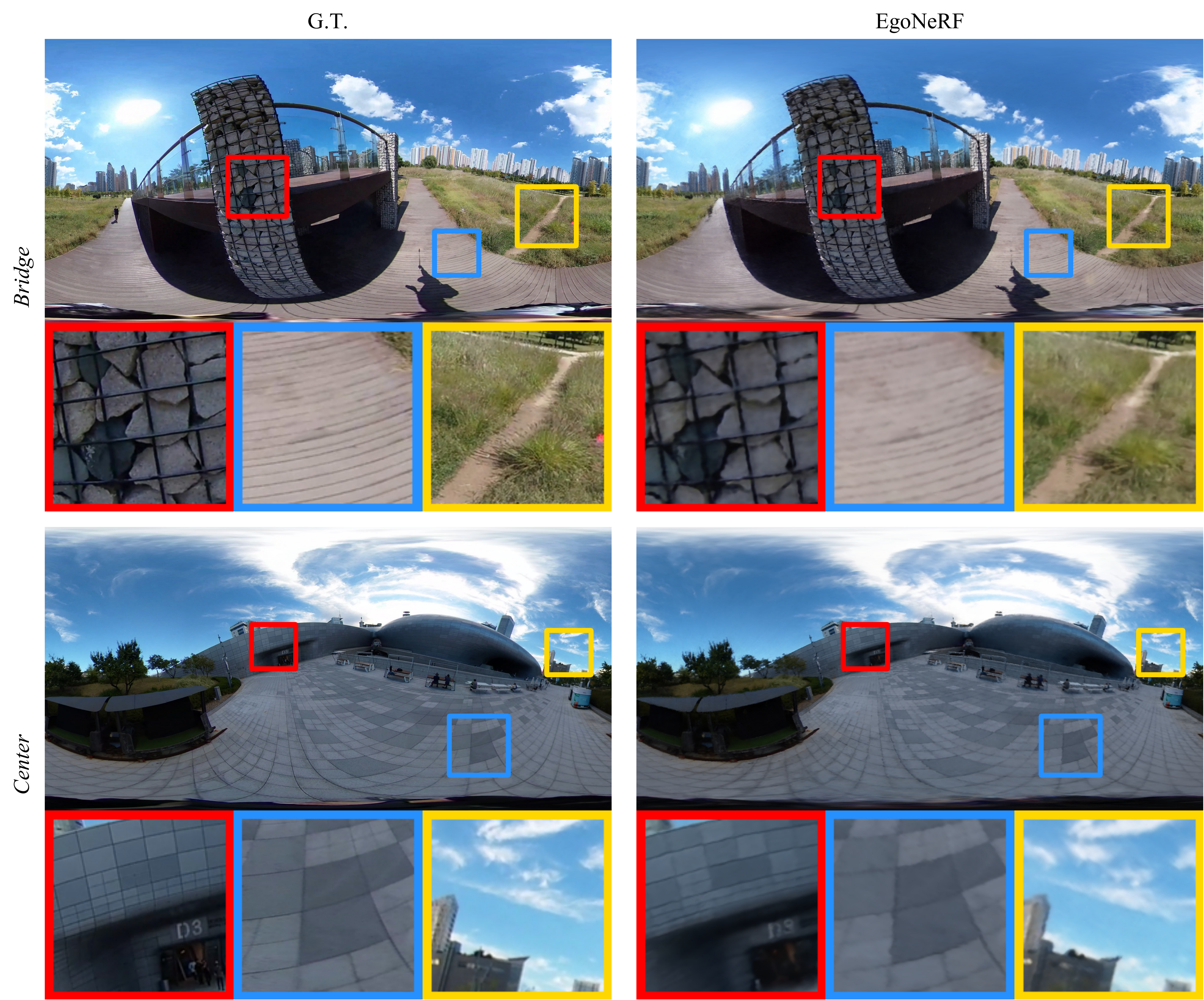}
\caption{Additional qualitative results in \textit{Ricoh360} dataset.}
\label{fig:additional_ricoh}
\end{figure*}

\Cref{fig:bistro_bike_geometry} illustrates the geometry obtained from our density feature grid $\mathcal{G}_\sigma$.
The explicit mesh is obtained by applying the marching cube algorithm~\cite{lorensen1987marching}.
Our approach is able to reconstruct fine details of large-scale scenes from a very short camera trajectory.
We also demonstrate additional rendered results on \textit{OmniBlender} and \textit{Ricoh360} datasets in \cref{fig:additional_omniblender} and \cref{fig:additional_ricoh}, respectively.
The results show that our approach is able to render high-quality images in both large-scale indoor and outdoor scenes from novel viewpoints.
We also provide per-scene breakdown for \textit{OmniBlender}, \textit{Ricoh360}, Synthetic-NeRF~\cite{mildenhall2021nerf}, and mip-NeRF 360~\cite{Barron_2022_CVPR} dataset in \cref{tab:omniblender_psnr,tab:omniblender_wspsnr,tab:omniblender_lpips,tab:omniblender_ssim,tab:omniblender_wsssim,tab:ricoh_psnr,tab:ricoh_wspsnr,tab:ricoh_lpips,tab:ricoh_ssim,tab:ricoh_wsssim,tab:synthetic_nerf,tab:mip-nerf_psnr,tab:mip-nerf_lpips,tab:mip-nerf_ssim}.

\begin{table*}
\centering
\resizebox{\linewidth}{!}{
\begin{tabular}{@{}l@{\:}l|cccc|ccccccc@{}}
\multirow{2}{*}{Step} & \multirow{2}{*}{Method} & \multicolumn{4}{c|}{Indoor} & \multicolumn{7}{c}{Outdoor} \\
& & \textit{BarberShop} & \textit{ClassRoom} & \textit{ItalianFlat} & \textit{Restroom} & \textit{BistroBike} & \textit{BistroSquare} & \textit{FisherHut} & \textit{LoneMonk} & \textit{LOU} & \textit{PavilionChair} & \textit{PavilionPond}\\
\hline
\multirow{5}{*}{5k} & NeRF~\cite{mildenhall2021nerf} & 26.32 & 24.70 & 26.31 & 27.67 & 20.31 & 17.80 & 26.91 & 21.95 & 23.57 & 25.33 & 20.65\\
 & mip-NeRF 360~\cite{Barron_2022_CVPR} & 22.65 & 21.23 & 24.33 & 25.83 & 20.79 & 18.68 & 26.99 & 20.38 & 21.98 & 23.72 & 19.77\\
 & TensoRF~\cite{chen2022tensorf} & 26.03 & 24.30 & 27.67 & 25.62 & 20.72 & 18.84 & 28.52 & 22.05 & 26.27 & 25.26 & 20.79\\
 & DVGO~\cite{Sun_2022_CVPR} & 22.84 & 22.31 & 25.24 & 26.65 & 19.24 & 17.87 & 27.56 & 20.30 & 23.45 & 24.14 & 19.34\\
 & EgoNeRF & \textbf{31.12} & \textbf{26.35} & \textbf{29.01} & \textbf{29.01} & \textbf{29.96} & \textbf{23.85} & \textbf{29.71} & \textbf{28.27} & \textbf{30.80} & \textbf{28.99} & \textbf{23.73}\\
\hline
\multirow{5}{*}{10k} & NeRF~\cite{mildenhall2021nerf} & 28.01 & 26.75 & 27.46 & 28.41 & 21.50 & 18.64 & 27.90 & 23.90 & 25.49 & 26.05 & 21.94\\
 & mip-NeRF 360~\cite{Barron_2022_CVPR} & 28.35 & 24.50 & 28.76 & 28.03 & 25.27 & 21.82 & 29.02 & 25.18 & 27.81 & 26.85 & 23.03\\
 & TensoRF~\cite{chen2022tensorf} & 27.54 & 25.22 & 28.81 & 26.26 & 21.74 & 19.49 & 28.85 & 22.96 & 28.04 & 26.07 & 21.47\\
 & DVGO~\cite{Sun_2022_CVPR} & 24.47 & 23.51 & 26.42 & 27.37 & 20.11 & 18.31 & 28.16 & 21.28 & 25.08 & 24.96 & 19.90\\
 & EgoNeRF & \textbf{32.53} & \textbf{27.47} & \textbf{30.48} & \textbf{30.43} & \textbf{31.29} & \textbf{24.52} & \textbf{30.01} & \textbf{29.28} & \textbf{32.01} & \textbf{29.86} & \textbf{24.68}\\
\hline
\multirow{5}{*}{100k} & NeRF~\cite{mildenhall2021nerf} & 33.20 & \textbf{31.05} & 30.17 & 32.24 & 25.29 & 20.85 & 30.10 & 28.23 & 31.43 & 29.28 & 24.68\\
 & mip-NeRF 360~\cite{Barron_2022_CVPR} & 33.30 & 26.83 & 32.82 & 31.54 & 30.62 & 24.93 & \textbf{31.13} & 30.22 & 32.59 & 30.53 & 25.39\\
 & TensoRF~\cite{chen2022tensorf} & 30.20 & 28.91 & 31.00 & 26.91 & 23.55 & 20.50 & 29.59 & 24.64 & 31.35 & 27.70 & 22.43\\
 & DVGO~\cite{Sun_2022_CVPR} & 29.21 & 26.70 & 30.14 & 29.29 & 22.69 & 19.87 & 29.54 & 23.50 & 29.75 & 27.24 & 21.48\\
 & EgoNeRF & \textbf{35.10} & 30.37 & \textbf{33.30} & \textbf{33.67} & \textbf{34.07} & \textbf{25.83} & 30.50 & \textbf{31.53} & \textbf{34.03} & \textbf{31.67} & \textbf{26.29}\\
\hline
\end{tabular}
}
\caption{Per-scene quantitative results in terms of PSNR in \textit{OmniBlender} dataset.}
\label{tab:omniblender_psnr}
\end{table*}

\begin{table*}
\centering
\resizebox{\linewidth}{!}{
\begin{tabular}{@{}l@{\:}l|cccc|ccccccc@{}}
\multirow{2}{*}{Step} & \multirow{2}{*}{Method} & \multicolumn{4}{c|}{Indoor} & \multicolumn{7}{c}{Outdoor} \\
& & \textit{BarberShop} & \textit{ClassRoom} & \textit{ItalianFlat} & \textit{Restroom} & \textit{BistroBike} & \textit{BistroSquare} & \textit{FisherHut} & \textit{LoneMonk} & \textit{LOU} & \textit{PavilionChair} & \textit{PavilionPond}\\
\hline
\multirow{5}{*}{5k} & NeRF~\cite{mildenhall2021nerf} & 27.09 & 25.38 & 27.30 & 29.32 & 21.54 & 19.69 & 27.96 & 23.45 & 24.49 & 26.10 & 22.08\\
 & mip-NeRF 360~\cite{Barron_2022_CVPR} & 23.40 & 21.71 & 25.07 & 27.46 & 21.80 & 20.74 & 28.10 & 21.89 & 22.74 & 24.78 & 21.13\\
 & TensoRF~\cite{chen2022tensorf} & 26.88 & 24.94 & 28.86 & 27.03 & 21.94 & 21.14 & 29.99 & 24.08 & 27.05 & 26.47 & 22.52\\
 & DVGO~\cite{Sun_2022_CVPR} & 23.66 & 22.94 & 26.14 & 28.40 & 20.40 & 19.98 & 28.83 & 21.98 & 24.20 & 25.34 & 21.29\\
 & EgoNeRF & \textbf{32.07} & \textbf{26.94} & \textbf{30.28} & \textbf{30.95} & \textbf{30.91} & \textbf{26.10} & \textbf{31.53} & \textbf{30.04} & \textbf{31.73} & \textbf{29.76} & \textbf{25.06}\\
\hline
\multirow{5}{*}{10k} & NeRF~\cite{mildenhall2021nerf} & 28.80 & 27.60 & 28.61 & 30.20 & 22.76 & 20.54 & 29.12 & 25.39 & 26.35 & 26.87 & 23.25\\
 & mip-NeRF 360~\cite{Barron_2022_CVPR} & 29.15 & 25.01 & 29.95 & 29.77 & 26.10 & 23.84 & 30.46 & 26.74 & 28.65 & 27.65 & 24.16\\
 & TensoRF~\cite{chen2022tensorf} & 28.40 & 25.86 & 30.19 & 27.49 & 23.07 & 21.89 & 30.42 & 25.23 & 28.90 & 27.25 & 23.21\\
 & DVGO~\cite{Sun_2022_CVPR} & 25.36 & 24.17 & 27.45 & 29.13 & 21.28 & 20.57 & 29.56 & 23.14 & 25.86 & 26.18 & 21.81\\
 & EgoNeRF & \textbf{33.49} & \textbf{28.11} & \textbf{31.92} & \textbf{32.36} & \textbf{32.22} & \textbf{26.88} & \textbf{31.84} & \textbf{31.12} & \textbf{32.96} & \textbf{30.60} & \textbf{25.86}\\
\hline
\multirow{5}{*}{100k} & NeRF~\cite{mildenhall2021nerf} & 34.13 & \textbf{32.29} & 31.86 & 34.02 & 26.87 & 22.98 & 31.77 & 29.92 & 32.41 & 29.99 & 25.81\\
 & mip-NeRF 360~\cite{Barron_2022_CVPR} & 34.20 & 27.37 & 34.59 & 33.48 & 31.39 & 27.24 & \textbf{32.93} & 31.86 & 33.49 & 31.13 & 26.38\\
 & TensoRF~\cite{chen2022tensorf} & 31.16 & 30.00 & 32.90 & 28.21 & 25.04 & 23.09 & 31.47 & 27.01 & 32.35 & 29.02 & 24.31\\
 & DVGO~\cite{Sun_2022_CVPR} & 30.25 & 27.49 & 31.85 & 31.32 & 24.23 & 22.51 & 31.32 & 26.02 & 30.73 & 28.63 & 23.64\\
 & EgoNeRF & \textbf{36.13} & 31.30 & \textbf{35.07} & \textbf{35.60} & \textbf{35.03} & \textbf{28.32} & 32.49 & \textbf{33.42} & \textbf{35.10} & \textbf{32.48} & \textbf{27.46}\\
\hline
\end{tabular}
}
\caption{Per-scene quantitative results in terms of $\text{PSNR}^{\text{WS}}$ in \textit{OmniBlender} dataset.}
\label{tab:omniblender_wspsnr}
\end{table*}

\begin{table*}
\centering
\resizebox{\linewidth}{!}{
\begin{tabular}{@{}l@{\:}l|cccc|ccccccc@{}}
\multirow{2}{*}{Step} & \multirow{2}{*}{Method} & \multicolumn{4}{c|}{Indoor} & \multicolumn{7}{c}{Outdoor} \\
& & \textit{BarberShop} & \textit{ClassRoom} & \textit{ItalianFlat} & \textit{Restroom} & \textit{BistroBike} & \textit{BistroSquare} & \textit{FisherHut} & \textit{LoneMonk} & \textit{LOU} & \textit{PavilionChair} & \textit{PavilionPond}\\
\hline
\multirow{5}{*}{5k} & NeRF~\cite{mildenhall2021nerf} & 0.420 & 0.560 & 0.373 & 0.646 & 0.658 & 0.711 & 0.503 & 0.448 & 0.418 & 0.368 & 0.561\\
 & mip-NeRF 360~\cite{Barron_2022_CVPR} & 0.613 & 0.662 & 0.470 & 0.768 & 0.624 & 0.573 & 0.462 & 0.531 & 0.495 & 0.536 & 0.596\\
 & TensoRF~\cite{chen2022tensorf} & 0.453 & 0.595 & 0.328 & 0.834 & 0.658 & 0.639 & 0.479 & 0.465 & 0.294 & 0.457 & 0.506\\
 & DVGO~\cite{Sun_2022_CVPR} & 0.605 & 0.714 & 0.425 & 0.788 & 0.721 & 0.711 & 0.499 & 0.544 & 0.400 & 0.529 & 0.584\\
 & EgoNeRF & \textbf{0.178} & \textbf{0.370} & \textbf{0.249} & \textbf{0.444} & \textbf{0.102} & \textbf{0.160} & \textbf{0.310} & \textbf{0.145} & \textbf{0.116} & \textbf{0.137} & \textbf{0.196}\\
\hline
\multirow{5}{*}{10k} & NeRF~\cite{mildenhall2021nerf} & 0.324 & 0.491 & 0.333 & 0.551 & 0.562 & 0.678 & 0.490 & 0.361 & 0.345 & 0.302 & 0.468\\
 & mip-NeRF 360~\cite{Barron_2022_CVPR} & 0.346 & 0.482 & 0.275 & 0.544 & 0.299 & 0.303 & 0.418 & 0.266 & 0.257 & 0.297 & 0.301\\
 & TensoRF~\cite{chen2022tensorf} & 0.345 & 0.517 & 0.274 & 0.738 & 0.574 & 0.566 & 0.467 & 0.409 & 0.228 & 0.374 & 0.437\\
 & DVGO~\cite{Sun_2022_CVPR} & 0.501 & 0.637 & 0.366 & 0.720 & 0.674 & 0.665 & 0.485 & 0.485 & 0.330 & 0.471 & 0.515\\
 & EgoNeRF & \textbf{0.128} & \textbf{0.323} & \textbf{0.189} & \textbf{0.350} & \textbf{0.074} & \textbf{0.126} & \textbf{0.281} & \textbf{0.115} & \textbf{0.095} & \textbf{0.099} & \textbf{0.164}\\
\hline
\multirow{5}{*}{100k} & NeRF~\cite{mildenhall2021nerf} & 0.133 & 0.324 & 0.234 & 0.269 & 0.278 & 0.496 & 0.357 & 0.166 & 0.148 & 0.139 & 0.301\\
 & mip-NeRF 360~\cite{Barron_2022_CVPR} & 0.135 & 0.321 & 0.134 & 0.308 & 0.092 & 0.113 & 0.261 & 0.107 & 0.121 & 0.099 & 0.150\\
 & TensoRF~\cite{chen2022tensorf} & 0.237 & 0.410 & 0.209 & 0.647 & 0.468 & 0.444 & 0.424 & 0.299 & 0.155 & 0.269 & 0.347\\
 & DVGO~\cite{Sun_2022_CVPR} & 0.248 & 0.426 & 0.220 & 0.498 & 0.487 & 0.508 & 0.433 & 0.325 & 0.180 & 0.281 & 0.325\\
 & EgoNeRF & \textbf{0.070} & \textbf{0.241} & \textbf{0.082} & \textbf{0.175} & \textbf{0.037} & \textbf{0.081} & \textbf{0.204} & \textbf{0.069} & \textbf{0.059} & \textbf{0.053} & \textbf{0.106}\\
\hline
\end{tabular}
}
\caption{Per-scene quantitative results in terms of LPIPS in \textit{OmniBlender} dataset.}
\label{tab:omniblender_lpips}
\end{table*}

\begin{table*}
\centering
\resizebox{\linewidth}{!}{
\begin{tabular}{@{}l@{\:}l|cccc|ccccccc@{}}
\multirow{2}{*}{Step} & \multirow{2}{*}{Method} & \multicolumn{4}{c|}{Indoor} & \multicolumn{7}{c}{Outdoor} \\
& & \textit{BarberShop} & \textit{ClassRoom} & \textit{ItalianFlat} & \textit{Restroom} & \textit{BistroBike} & \textit{BistroSquare} & \textit{FisherHut} & \textit{LoneMonk} & \textit{LOU} & \textit{PavilionChair} & \textit{PavilionPond}\\
\hline
\multirow{5}{*}{5k} & NeRF~\cite{mildenhall2021nerf} & 0.801 & 0.698 & 0.804 & 0.603 & 0.554 & 0.517 & 0.737 & 0.644 & 0.742 & 0.777 & 0.587\\
 & mip-NeRF 360~\cite{Barron_2022_CVPR} & 0.685 & 0.613 & 0.752 & 0.545 & 0.539 & 0.528 & 0.737 & 0.570 & 0.678 & 0.710 & 0.538\\
 & TensoRF~\cite{chen2022tensorf} & 0.796 & 0.702 & 0.828 & 0.563 & 0.567 & 0.551 & 0.754 & 0.653 & 0.827 & 0.761 & 0.593\\
 & DVGO~\cite{Sun_2022_CVPR} & 0.733 & 0.655 & 0.796 & 0.572 & 0.536 & 0.523 & 0.744 & 0.611 & 0.752 & 0.745 & 0.582\\
 & EgoNeRF & \textbf{0.905} & \textbf{0.766} & \textbf{0.849} & \textbf{0.694} & \textbf{0.906} & \textbf{0.829} & \textbf{0.777} & \textbf{0.875} & \textbf{0.898} & \textbf{0.881} & \textbf{0.738}\\
\hline
\multirow{5}{*}{10k} & NeRF~\cite{mildenhall2021nerf} & 0.838 & 0.732 & 0.820 & 0.636 & 0.594 & 0.532 & 0.747 & 0.714 & 0.786 & 0.800 & 0.627\\
 & mip-NeRF 360~\cite{Barron_2022_CVPR} & 0.845 & 0.724 & 0.848 & 0.637 & 0.764 & 0.723 & 0.768 & 0.777 & 0.852 & 0.809 & 0.686\\
 & TensoRF~\cite{chen2022tensorf} & 0.839 & 0.730 & 0.845 & 0.592 & 0.605 & 0.576 & 0.759 & 0.684 & 0.867 & 0.776 & 0.609\\
 & DVGO~\cite{Sun_2022_CVPR} & 0.769 & 0.682 & 0.813 & 0.595 & 0.553 & 0.534 & 0.751 & 0.637 & 0.791 & 0.757 & 0.592\\
 & EgoNeRF & \textbf{0.930} & \textbf{0.794} & \textbf{0.876} & \textbf{0.761} & \textbf{0.930} & \textbf{0.862} & \textbf{0.788} & \textbf{0.901} & \textbf{0.914} & \textbf{0.905} & \textbf{0.774}\\
\hline
\multirow{5}{*}{100k} & NeRF~\cite{mildenhall2021nerf} & 0.927 & 0.813 & 0.858 & 0.809 & 0.761 & 0.608 & 0.779 & 0.860 & 0.894 & 0.885 & 0.736\\
 & mip-NeRF 360~\cite{Barron_2022_CVPR} & 0.937 & 0.803 & 0.912 & 0.783 & 0.924 & 0.881 & \textbf{0.813} & 0.910 & 0.924 & 0.913 & 0.790\\
 & TensoRF~\cite{chen2022tensorf} & 0.887 & 0.782 & 0.871 & 0.624 & 0.668 & 0.608 & 0.770 & 0.735 & 0.906 & 0.810 & 0.641\\
 & DVGO~\cite{Sun_2022_CVPR} & 0.874 & 0.765 & 0.863 & 0.688 & 0.646 & 0.599 & 0.769 & 0.711 & 0.873 & 0.802 & 0.635\\
 & EgoNeRF & \textbf{0.960} & \textbf{0.843} & \textbf{0.930} & \textbf{0.873} & \textbf{0.962} & \textbf{0.909} & 0.811 & \textbf{0.940} & \textbf{0.935} & \textbf{0.941} & \textbf{0.831}\\
\hline
\end{tabular}
}
\caption{Per-scene quantitative results in terms of SSIM in \textit{OmniBlender} dataset.}
\label{tab:omniblender_ssim}
\end{table*}

\begin{table*}
\centering
\resizebox{\linewidth}{!}{
\begin{tabular}{@{}l@{\:}l|cccc|ccccccc@{}}
\multirow{2}{*}{Step} & \multirow{2}{*}{Method} & \multicolumn{4}{c|}{Indoor} & \multicolumn{7}{c}{Outdoor} \\
& & \textit{BarberShop} & \textit{ClassRoom} & \textit{ItalianFlat} & \textit{Restroom} & \textit{BistroBike} & \textit{BistroSquare} & \textit{FisherHut} & \textit{LoneMonk} & \textit{LOU} & \textit{PavilionChair} & \textit{PavilionPond}\\
\hline
\multirow{5}{*}{5k} & NeRF~\cite{mildenhall2021nerf} & 0.763 & 0.682 & 0.796 & 0.601 & 0.487 & 0.479 & 0.712 & 0.614 & 0.695 & 0.728 & 0.561\\
 & mip-NeRF 360~\cite{Barron_2022_CVPR} & 0.616 & 0.573 & 0.725 & 0.540 & 0.468 & 0.494 & 0.712 & 0.536 & 0.609 & 0.659 & 0.502\\
 & TensoRF~\cite{chen2022tensorf} & 0.759 & 0.689 & 0.829 & 0.556 & 0.503 & 0.530 & 0.737 & 0.650 & 0.797 & 0.720 & 0.576\\
 & DVGO~\cite{Sun_2022_CVPR} & 0.677 & 0.627 & 0.785 & 0.574 & 0.462 & 0.492 & 0.722 & 0.592 & 0.709 & 0.697 & 0.559\\
 & EgoNeRF & \textbf{0.895} & \textbf{0.760} & \textbf{0.852} & \textbf{0.704} & \textbf{0.894} & \textbf{0.824} & \textbf{0.772} & \textbf{0.871} & \textbf{0.888} & \textbf{0.853} & \textbf{0.722}\\
\hline
\multirow{5}{*}{10k} & NeRF~\cite{mildenhall2021nerf} & 0.809 & 0.726 & 0.818 & 0.641 & 0.538 & 0.495 & 0.725 & 0.687 & 0.746 & 0.756 & 0.600\\
 & mip-NeRF 360~\cite{Barron_2022_CVPR} & 0.820 & 0.709 & 0.852 & 0.638 & 0.720 & 0.701 & 0.755 & 0.759 & 0.828 & 0.768 & 0.657\\
 & TensoRF~\cite{chen2022tensorf} & 0.814 & 0.724 & 0.852 & 0.581 & 0.552 & 0.562 & 0.744 & 0.693 & 0.846 & 0.740 & 0.598\\
 & DVGO~\cite{Sun_2022_CVPR} & 0.725 & 0.664 & 0.808 & 0.598 & 0.484 & 0.509 & 0.732 & 0.629 & 0.759 & 0.714 & 0.574\\
 & EgoNeRF & \textbf{0.925} & \textbf{0.792} & \textbf{0.883} & \textbf{0.764} & \textbf{0.922} & \textbf{0.861} & \textbf{0.784} & \textbf{0.899} & \textbf{0.907} & \textbf{0.882} & \textbf{0.757}\\
\hline
\multirow{5}{*}{100k} & NeRF~\cite{mildenhall2021nerf} & 0.922 & 0.821 & 0.868 & 0.803 & 0.746 & 0.588 & 0.768 & 0.849 & 0.882 & 0.856 & 0.712\\
 & mip-NeRF 360~\cite{Barron_2022_CVPR} & 0.934 & 0.801 & 0.919 & 0.782 & 0.912 & 0.876 & 0.805 & 0.902 & 0.917 & 0.887 & 0.767\\
 & TensoRF~\cite{chen2022tensorf} & 0.875 & 0.793 & 0.885 & 0.619 & 0.634 & 0.608 & 0.762 & 0.748 & 0.896 & 0.787 & 0.643\\
 & DVGO~\cite{Sun_2022_CVPR} & 0.862 & 0.772 & 0.875 & 0.703 & 0.608 & 0.596 & 0.759 & 0.729 & 0.867 & 0.778 & 0.641\\
 & EgoNeRF & \textbf{0.960} & \textbf{0.848} & \textbf{0.937} & \textbf{0.870} & \textbf{0.959} & \textbf{0.913} & \textbf{0.813} & \textbf{0.939} & \textbf{0.933} & \textbf{0.927} & \textbf{0.823}\\
\hline
\end{tabular}
}
\caption{Per-scene quantitative results in terms of $\text{SSIM}^{\text{WS}}$ in \textit{OmniBlender} dataset.}
\label{tab:omniblender_wsssim}
\end{table*}

\begin{table*}
\centering
\resizebox{\linewidth}{!}{
\begin{tabular}{@{}l@{\:}l|ccccccccccc@{}}
Step & Method & \textit{Bricks} & \textit{Bridge} & \textit{BridgeUnder} & \textit{CatTower} & \textit{Center} & \textit{Farm} & \textit{Flower} & \textit{GalleryChair} & \textit{GalleryPillar} & \textit{Garden} & \textit{Poster}\\
\hline
\multirow{5}{*}{5k} & NeRF~\cite{mildenhall2021nerf} & 20.05 & 20.91 & 21.71 & 21.65 & 24.75 & 19.84 & 18.91 & 24.75 & 24.34 & 24.06 & 22.03\\
 & mip-NeRF 360~\cite{Barron_2022_CVPR} & 20.22 & 20.87 & 20.87 & 22.15 & 25.53 & 20.07 & 19.25 & 24.96 & 24.90 & 25.13 & 21.34\\
 & TensoRF~\cite{chen2022tensorf} & 20.97 & 21.75 & 21.92 & 22.39 & 26.91 & 20.89 & 20.09 & 25.91 & 26.02 & 25.13 & 23.20\\
 & DVGO~\cite{Sun_2022_CVPR} & 20.19 & 20.91 & 20.70 & 21.77 & 26.15 & 20.33 & 19.60 & 25.27 & 25.20 & 25.00 & 21.82\\
 & EgoNeRF & \textbf{22.44} & \textbf{22.86} & \textbf{23.99} & \textbf{23.49} & \textbf{27.88} & \textbf{21.83} & \textbf{21.31} & \textbf{26.98} & \textbf{27.29} & \textbf{26.31} & \textbf{25.29}\\
\hline
\multirow{5}{*}{10k} & NeRF~\cite{mildenhall2021nerf} & 20.64 & 21.48 & 22.43 & 22.18 & 25.81 & 20.29 & 19.52 & 25.60 & 25.30 & 24.49 & 22.79\\
 & mip-NeRF 360~\cite{Barron_2022_CVPR} & 22.08 & 22.73 & 23.37 & 23.38 & 27.73 & 21.66 & 20.93 & 27.03 & 26.97 & 26.09 & 25.11\\
 & TensoRF~\cite{chen2022tensorf} & 21.62 & 22.17 & 22.78 & 22.80 & 27.61 & 21.20 & 20.63 & 26.68 & 26.60 & 25.57 & 24.31\\
 & DVGO~\cite{Sun_2022_CVPR} & 20.84 & 21.64 & 21.43 & 22.29 & 26.92 & 20.62 & 20.12 & 25.55 & 26.16 & 25.24 & 23.07\\
 & EgoNeRF & \textbf{22.68} & \textbf{22.98} & \textbf{24.25} & \textbf{23.69} & \textbf{28.07} & \textbf{21.98} & \textbf{21.51} & \textbf{27.13} & \textbf{27.50} & \textbf{26.50} & \textbf{25.57}\\
\hline
\multirow{5}{*}{100k} & NeRF~\cite{mildenhall2021nerf} & 22.71 & 23.39 & 24.82 & 23.99 & 28.54 & 21.68 & 21.44 & 27.77 & 27.43 & 26.42 & 25.85\\
 & mip-NeRF 360~\cite{Barron_2022_CVPR} & \textbf{23.39} & \textbf{23.80} & \textbf{25.01} & \textbf{24.46} & \textbf{29.30} & \textbf{22.48} & \textbf{22.01} & \textbf{28.36} & \textbf{28.42} & \textbf{27.03} & \textbf{27.00}\\
 & TensoRF~\cite{chen2022tensorf} & 23.08 & 23.27 & 24.56 & 23.84 & 29.25 & 22.02 & 21.72 & 28.04 & 28.14 & 26.47 & 26.38\\
 & DVGO~\cite{Sun_2022_CVPR} & 22.97 & 22.94 & 23.96 & 23.84 & 29.21 & 21.79 & 21.72 & 26.49 & 28.28 & 26.40 & 26.24\\
 & EgoNeRF & 23.37 & 23.40 & 24.94 & 24.23 & 28.45 & 22.23 & 21.80 & 27.78 & 28.02 & 26.87 & 26.62\\
\hline
\end{tabular}
}
\caption{Per-scene quantitative results in terms of PSNR in \textit{Ricoh360} dataset.}
\label{tab:ricoh_psnr}
\end{table*}

\begin{table*}
\centering
\resizebox{\linewidth}{!}{
\begin{tabular}{@{}l@{\:}l|ccccccccccc@{}}
Step & Method & \textit{Bricks} & \textit{Bridge} & \textit{BridgeUnder} & \textit{CatTower} & \textit{Center} & \textit{Farm} & \textit{Flower} & \textit{GalleryChair} & \textit{GalleryPillar} & \textit{Garden} & \textit{Poster}\\
\hline
\multirow{5}{*}{5k} & NeRF~\cite{mildenhall2021nerf} & 21.79 & 22.68 & 23.15 & 23.47 & 26.71 & 21.52 & 20.73 & 27.14 & 25.78 & 25.62 & 23.41\\
 & mip-NeRF 360~\cite{Barron_2022_CVPR} & 22.03 & 22.68 & 22.60 & 24.09 & 27.54 & 21.88 & 21.21 & 27.39 & 26.35 & 26.75 & 22.77\\
 & TensoRF~\cite{chen2022tensorf} & 22.95 & 23.88 & 23.71 & 24.45 & 28.91 & 22.78 & 22.10 & 28.53 & 27.65 & 26.84 & 24.95\\
 & DVGO~\cite{Sun_2022_CVPR} & 22.47 & 23.23 & 22.74 & 23.89 & 28.30 & 22.43 & 21.74 & 28.31 & 26.99 & 26.86 & 23.49\\
 & EgoNeRF & \textbf{24.81} & \textbf{25.25} & \textbf{25.95} & \textbf{25.63} & \textbf{30.39} & \textbf{23.97} & \textbf{23.44} & \textbf{29.89} & \textbf{29.19} & \textbf{28.01} & \textbf{27.61}\\
\hline
\multirow{5}{*}{10k} & NeRF~\cite{mildenhall2021nerf} & 22.35 & 23.30 & 23.87 & 24.05 & 27.60 & 22.00 & 21.39 & 27.86 & 26.75 & 26.04 & 24.21\\
 & mip-NeRF 360~\cite{Barron_2022_CVPR} & 24.10 & 24.80 & 25.20 & 25.36 & 30.15 & 23.59 & 22.91 & 29.58 & 28.69 & 27.65 & 27.09\\
 & TensoRF~\cite{chen2022tensorf} & 23.59 & 24.33 & 24.50 & 24.87 & 29.59 & 23.07 & 22.61 & 29.18 & 28.17 & 27.19 & 26.17\\
 & DVGO~\cite{Sun_2022_CVPR} & 23.04 & 23.95 & 23.43 & 24.45 & 29.22 & 22.79 & 22.31 & 28.84 & 27.99 & 27.19 & 24.82\\
 & EgoNeRF & \textbf{25.08} & \textbf{25.46} & \textbf{26.26} & \textbf{25.84} & \textbf{30.65} & \textbf{24.16} & \textbf{23.67} & \textbf{30.07} & \textbf{29.43} & \textbf{28.19} & \textbf{27.93}\\
\hline
\multirow{5}{*}{100k} & NeRF~\cite{mildenhall2021nerf} & 24.42 & 25.09 & 26.38 & 25.90 & 30.27 & 23.40 & 23.38 & 30.04 & 29.01 & 27.87 & 27.37\\
 & mip-NeRF 360~\cite{Barron_2022_CVPR} & 25.52 & \textbf{25.96} & 26.90 & \textbf{26.48} & \textbf{31.52} & \textbf{24.50} & \textbf{24.03} & \textbf{30.89} & \textbf{30.20} & \textbf{28.66} & \textbf{29.13}\\
 & TensoRF~\cite{chen2022tensorf} & 25.06 & 25.42 & 26.26 & 26.00 & 31.12 & 23.97 & 23.72 & 30.56 & 29.81 & 28.08 & 28.46\\
 & DVGO~\cite{Sun_2022_CVPR} & 25.23 & 25.41 & 25.91 & 26.20 & 31.42 & 24.00 & 24.00 & 30.62 & 30.36 & 28.38 & 28.55\\
 & EgoNeRF & \textbf{25.74} & 25.88 & \textbf{27.01} & 26.41 & 31.08 & 24.40 & 24.01 & 30.64 & 30.07 & 28.53 & 28.70\\
\hline
\end{tabular}
}
\caption{Per-scene quantitative results in terms of $\text{PSNR}^{\text{WS}}$ in \textit{Ricoh360} dataset.}
\label{tab:ricoh_wspsnr}
\end{table*}

\begin{table*}
\centering
\resizebox{\linewidth}{!}{
\begin{tabular}{@{}l@{\:}l|ccccccccccc@{}}
Step & Method & \textit{Bricks} & \textit{Bridge} & \textit{BridgeUnder} & \textit{CatTower} & \textit{Center} & \textit{Farm} & \textit{Flower} & \textit{GalleryChair} & \textit{GalleryPillar} & \textit{Garden} & \textit{Poster}\\
\hline
\multirow{5}{*}{5k} & NeRF~\cite{mildenhall2021nerf} & 0.587 & 0.542 & 0.559 & 0.631 & 0.516 & 0.599 & 0.725 & 0.572 & 0.460 & 0.577 & 0.566\\
 & mip-NeRF 360~\cite{Barron_2022_CVPR} & 0.569 & 0.524 & 0.629 & 0.575 & 0.472 & 0.583 & 0.662 & 0.518 & 0.438 & 0.531 & 0.599\\
 & TensoRF~\cite{chen2022tensorf} & 0.531 & 0.509 & 0.591 & 0.617 & 0.457 & 0.537 & 0.709 & 0.546 & 0.408 & 0.562 & 0.496\\
 & DVGO~\cite{Sun_2022_CVPR} & 0.547 & 0.540 & 0.680 & 0.636 & 0.473 & 0.574 & 0.714 & 0.552 & 0.456 & 0.574 & 0.552\\
 & EgoNeRF & \textbf{0.317} & \textbf{0.330} & \textbf{0.309} & \textbf{0.395} & \textbf{0.251} & \textbf{0.342} & \textbf{0.434} & \textbf{0.336} & \textbf{0.238} & \textbf{0.371} & \textbf{0.323}\\
\hline
\multirow{5}{*}{10k} & NeRF~\cite{mildenhall2021nerf} & 0.547 & 0.505 & 0.499 & 0.610 & 0.484 & 0.554 & 0.698 & 0.538 & 0.414 & 0.562 & 0.509\\
 & mip-NeRF 360~\cite{Barron_2022_CVPR} & 0.371 & 0.363 & 0.390 & 0.460 & 0.293 & 0.366 & 0.517 & 0.401 & 0.275 & 0.427 & 0.357\\
 & TensoRF~\cite{chen2022tensorf} & 0.469 & 0.459 & 0.484 & 0.575 & 0.387 & 0.484 & 0.653 & 0.485 & 0.349 & 0.529 & 0.415\\
 & DVGO~\cite{Sun_2022_CVPR} & 0.499 & 0.495 & 0.613 & 0.606 & 0.437 & 0.531 & 0.678 & 0.522 & 0.411 & 0.553 & 0.478\\
 & EgoNeRF & \textbf{0.292} & \textbf{0.312} & \textbf{0.282} & \textbf{0.380} & \textbf{0.236} & \textbf{0.322} & \textbf{0.424} & \textbf{0.323} & \textbf{0.227} & \textbf{0.361} & \textbf{0.290}\\
\hline
\multirow{5}{*}{100k} & NeRF~\cite{mildenhall2021nerf} & 0.396 & 0.378 & 0.292 & 0.472 & 0.295 & 0.404 & 0.540 & 0.383 & 0.288 & 0.424 & 0.350\\
 & mip-NeRF 360~\cite{Barron_2022_CVPR} & \textbf{0.246} & \textbf{0.258} & 0.238 & \textbf{0.337} & \textbf{0.192} & \textbf{0.265} & \textbf{0.378} & 0.301 & \textbf{0.180} & \textbf{0.312} & \textbf{0.239}\\
 & TensoRF~\cite{chen2022tensorf} & 0.342 & 0.360 & 0.332 & 0.487 & 0.279 & 0.378 & 0.530 & 0.385 & 0.274 & 0.457 & 0.314\\
 & DVGO~\cite{Sun_2022_CVPR} & 0.331 & 0.355 & 0.365 & 0.481 & 0.302 & 0.375 & 0.512 & 0.387 & 0.271 & 0.458 & 0.302\\
 & EgoNeRF & 0.254 & 0.276 & \textbf{0.231} & 0.369 & 0.207 & 0.312 & 0.412 & \textbf{0.288} & 0.206 & 0.342 & 0.245\\
\hline
\end{tabular}
}
\caption{Per-scene quantitative results in terms of LPIPS in \textit{Ricoh360} dataset.}
\label{tab:ricoh_lpips}
\end{table*}

\begin{table*}
\centering
\resizebox{\linewidth}{!}{
\begin{tabular}{@{}l@{\:}l|ccccccccccc@{}}
Step & Method & \textit{Bricks} & \textit{Bridge} & \textit{BridgeUnder} & \textit{CatTower} & \textit{Center} & \textit{Farm} & \textit{Flower} & \textit{GalleryChair} & \textit{GalleryPillar} & \textit{Garden} & \textit{Poster}\\
\hline
\multirow{5}{*}{5k} & NeRF~\cite{mildenhall2021nerf} & 0.577 & 0.624 & 0.621 & 0.606 & 0.768 & 0.540 & 0.512 & 0.774 & 0.755 & 0.646 & 0.722\\
 & mip-NeRF 360~\cite{Barron_2022_CVPR} & 0.563 & 0.605 & 0.574 & 0.617 & 0.748 & 0.507 & 0.516 & 0.758 & 0.737 & 0.657 & 0.667\\
 & TensoRF~\cite{chen2022tensorf} & 0.617 & 0.647 & 0.631 & 0.628 & 0.803 & 0.571 & 0.542 & 0.794 & 0.785 & 0.664 & 0.758\\
 & DVGO~\cite{Sun_2022_CVPR} & 0.609 & 0.637 & 0.598 & 0.620 & 0.797 & 0.564 & 0.532 & 0.784 & 0.773 & 0.660 & 0.732\\
 & EgoNeRF & \textbf{0.707} & \textbf{0.704} & \textbf{0.746} & \textbf{0.673} & \textbf{0.844} & \textbf{0.641} & \textbf{0.606} & \textbf{0.829} & \textbf{0.830} & \textbf{0.706} & \textbf{0.818}\\
\hline
\multirow{5}{*}{10k} & NeRF~\cite{mildenhall2021nerf} & 0.594 & 0.634 & 0.650 & 0.615 & 0.783 & 0.549 & 0.523 & 0.783 & 0.769 & 0.653 & 0.742\\
 & mip-NeRF 360~\cite{Barron_2022_CVPR} & 0.676 & 0.695 & 0.723 & 0.668 & 0.838 & 0.626 & 0.593 & 0.823 & 0.821 & 0.695 & 0.816\\
 & TensoRF~\cite{chen2022tensorf} & 0.639 & 0.657 & 0.667 & 0.640 & 0.819 & 0.585 & 0.557 & 0.806 & 0.800 & 0.672 & 0.787\\
 & DVGO~\cite{Sun_2022_CVPR} & 0.628 & 0.650 & 0.616 & 0.631 & 0.809 & 0.574 & 0.545 & 0.793 & 0.790 & 0.665 & 0.762\\
 & EgoNeRF & \textbf{0.720} & \textbf{0.713} & \textbf{0.763} & \textbf{0.681} & \textbf{0.850} & \textbf{0.651} & \textbf{0.617} & \textbf{0.834} & \textbf{0.835} & \textbf{0.713} & \textbf{0.831}\\
\hline
\multirow{5}{*}{100k} & NeRF~\cite{mildenhall2021nerf} & 0.670 & 0.687 & 0.755 & 0.659 & 0.830 & 0.607 & 0.579 & 0.825 & 0.815 & 0.690 & 0.815\\
 & mip-NeRF 360~\cite{Barron_2022_CVPR} & \textbf{0.761} & \textbf{0.748} & \textbf{0.801} & \textbf{0.713} & \textbf{0.872} & \textbf{0.690} & \textbf{0.651} & \textbf{0.859} & \textbf{0.860} & \textbf{0.739} & \textbf{0.866}\\
 & TensoRF~\cite{chen2022tensorf} & 0.701 & 0.695 & 0.736 & 0.665 & 0.849 & 0.631 & 0.595 & 0.831 & 0.831 & 0.692 & 0.832\\
 & DVGO~\cite{Sun_2022_CVPR} & 0.708 & 0.696 & 0.717 & 0.670 & 0.849 & 0.628 & 0.601 & 0.826 & 0.833 & 0.689 & 0.832\\
 & EgoNeRF & 0.748 & 0.733 & 0.791 & 0.697 & 0.858 & 0.665 & 0.633 & 0.847 & 0.846 & 0.725 & 0.853\\
\hline
\end{tabular}
}
\caption{Per-scene quantitative results in terms of SSIM in \textit{Ricoh360} dataset.}
\label{tab:ricoh_ssim}
\end{table*}

\begin{table*}
\centering
\resizebox{\linewidth}{!}{
\begin{tabular}{@{}l@{\:}l|ccccccccccc@{}}
Step & Method & \textit{Bricks} & \textit{Bridge} & \textit{BridgeUnder} & \textit{CatTower} & \textit{Center} & \textit{Farm} & \textit{Flower} & \textit{GalleryChair} & \textit{GalleryPillar} & \textit{Garden} & \textit{Poster}\\
\hline
\multirow{5}{*}{5k} & NeRF~\cite{mildenhall2021nerf} & 0.553 & 0.583 & 0.579 & 0.587 & 0.754 & 0.496 & 0.499 & 0.766 & 0.729 & 0.613 & 0.697\\
 & mip-NeRF 360~\cite{Barron_2022_CVPR} & 0.541 & 0.565 & 0.533 & 0.598 & 0.730 & 0.467 & 0.503 & 0.746 & 0.706 & 0.625 & 0.635\\
 & TensoRF~\cite{chen2022tensorf} & 0.604 & 0.614 & 0.600 & 0.616 & 0.794 & 0.534 & 0.535 & 0.789 & 0.767 & 0.635 & 0.747\\
 & DVGO~\cite{Sun_2022_CVPR} & 0.601 & 0.605 & 0.568 & 0.607 & 0.788 & 0.528 & 0.528 & 0.782 & 0.755 & 0.631 & 0.717\\
 & EgoNeRF & \textbf{0.704} & \textbf{0.681} & \textbf{0.732} & \textbf{0.668} & \textbf{0.842} & \textbf{0.618} & \textbf{0.608} & \textbf{0.831} & \textbf{0.824} & \textbf{0.683} & \textbf{0.824}\\
\hline
\multirow{5}{*}{10k} & NeRF~\cite{mildenhall2021nerf} & 0.571 & 0.595 & 0.611 & 0.596 & 0.770 & 0.506 & 0.511 & 0.775 & 0.744 & 0.621 & 0.721\\
 & mip-NeRF 360~\cite{Barron_2022_CVPR} & 0.665 & 0.665 & 0.700 & 0.657 & 0.832 & 0.594 & 0.588 & 0.820 & 0.808 & 0.668 & 0.814\\
 & TensoRF~\cite{chen2022tensorf} & 0.628 & 0.627 & 0.638 & 0.630 & 0.811 & 0.550 & 0.553 & 0.804 & 0.785 & 0.644 & 0.783\\
 & DVGO~\cite{Sun_2022_CVPR} & 0.623 & 0.622 & 0.590 & 0.621 & 0.802 & 0.542 & 0.545 & 0.794 & 0.776 & 0.640 & 0.755\\
 & EgoNeRF & \textbf{0.718} & \textbf{0.692} & \textbf{0.750} & \textbf{0.677} & \textbf{0.849} & \textbf{0.630} & \textbf{0.620} & \textbf{0.837} & \textbf{0.831} & \textbf{0.691} & \textbf{0.840}\\
\hline
\multirow{5}{*}{100k} & NeRF~\cite{mildenhall2021nerf} & 0.649 & 0.649 & 0.729 & 0.643 & 0.821 & 0.569 & 0.571 & 0.820 & 0.801 & 0.660 & 0.807\\
 & mip-NeRF 360~\cite{Barron_2022_CVPR} & \textbf{0.754} & \textbf{0.727} & \textbf{0.788} & \textbf{0.706} & \textbf{0.869} & \textbf{0.668} & \textbf{0.651} & \textbf{0.859} & \textbf{0.857} & \textbf{0.716} & \textbf{0.873}\\
 & TensoRF~\cite{chen2022tensorf} & 0.698 & 0.672 & 0.717 & 0.662 & 0.845 & 0.606 & 0.595 & 0.832 & 0.824 & 0.669 & 0.839\\
 & DVGO~\cite{Sun_2022_CVPR} & 0.715 & 0.679 & 0.705 & 0.672 & 0.849 & 0.609 & 0.610 & 0.835 & 0.833 & 0.671 & 0.844\\
 & EgoNeRF & 0.747 & 0.713 & 0.782 & 0.693 & 0.860 & 0.645 & 0.637 & 0.850 & 0.844 & 0.704 & 0.861\\
\hline
\end{tabular}
}
\caption{Per-scene quantitative results in terms of $\text{SSIM}^{\text{WS}}$ in \textit{Ricoh360} dataset.}
\label{tab:ricoh_wsssim}
\end{table*}

\begin{table*}
    \centering
    \resizebox{0.6\linewidth}{!}{
    \begin{tabular}{l|cccccccc}
    & \textit{Chair} & \textit{Drums} & \textit{Ficus} & \textit{Hotdog} & \textit{Lego} & \textit{Materials} & \textit{Mic} & \textit{Ship}\\
    \hline
    PSNR & 34.15 & 25.30 & 31.72 & 36.19 & 33.88 & 28.87 & 33.09 & 28.85\\
    LPIPS & 0.017 & 0.062 & 0.020 & 0.020 & 0.014 & 0.037 & 0.015 & 0.111\\
    SSIM & 0.977 & 0.928 & 0.972 & 0.978 & 0.972 & 0.941 & 0.982 & 0.862\\
    \end{tabular}
    }
    \caption{Per-scene breakdown of EgoNeRF results for Synthetic-NeRF~\cite{mildenhall2021nerf} dataset.}
    \label{tab:synthetic_nerf}
\end{table*}
\begin{table*}
\centering
\resizebox{0.7\linewidth}{!}{
\begin{tabular}{l|cccccccc}
Method & \textit{Bicycle} & \textit{Bonsai} & \textit{Counter} & \textit{Garden} & \textit{Kitchen} & \textit{Room} & \textit{Stump} \\
\hline
NeRF~\cite{mildenhall2021nerf} & 21.76 & 26.81 & 25.67 & 23.11 & 26.31 & 28.56 & 21.73\\
mip-NeRF ~\cite{Barron_2021_ICCV} & 21.69 & 27.13 & 25.59 & 23.16 & 26.47 & 28.73 & 23.10\\
mip-NeRF 360~\cite{Barron_2022_CVPR} & 24.37 & 33.46 & 29.55 & 26.98 & 32.23 & 31.63 & 26.40\\
TensoRF~\cite{chen2022tensorf} & 19.86 & 24.99 & 22.58 & 21.96 & 22.78 & 26.13 & 20.91\\
DVGO~\cite{Sun_2022_CVPR} & 18.88 & 16.89 & 22.54 & 19.39 & 21.63 & 26.89 & 20.40\\
EgoNeRF & 21.88 & 28.62 & 25.82 & 24.41 & 26.48 & 29.68 & 23.94\\
\end{tabular}
}
\caption{Per-scene quantitative results in terms of PSNR in mip-NeRF 360~\cite{Barron_2022_CVPR} dataset.}
\label{tab:mip-nerf_psnr}
\end{table*}

\begin{table*}
\centering
\resizebox{0.7\linewidth}{!}{
\begin{tabular}{l|cccccccc}
Method & \textit{Bicycle} & \textit{Bonsai} & \textit{Counter} & \textit{Garden} & \textit{Kitchen} & \textit{Room} & \textit{Stump} \\
\hline
NeRF~\cite{mildenhall2021nerf} & 0.536 & 0.398 & 0.394 & 0.415 & 0.335 & 0.353 & 0.551\\
mip-NeRF ~\cite{Barron_2021_ICCV} & 0.541 & 0.370 & 0.390 & 0.422 & 0.336 & 0.346 & 0.490\\
mip-NeRF 360~\cite{Barron_2022_CVPR} & 0.301 & 0.176 & 0.204 & 0.170 & 0.127 & 0.211 & 0.261\\
TensoRF~\cite{chen2022tensorf} & 0.838 & 0.414 & 0.578 & 0.728 & 0.578 & 0.456 & 0.738\\
DVGO~\cite{Sun_2022_CVPR} & 0.687 & 0.639 & 0.405 & 0.529 & 0.430 & 0.331 & 0.589\\
EgoNeRF & 0.507 & 0.220 & 0.319 & 0.318 & 0.250 & 0.273 & 0.357\\
\end{tabular}
}
\caption{Per-scene quantitative results in terms of LPIPS in mip-NeRF 360~\cite{Barron_2022_CVPR} dataset.}
\label{tab:mip-nerf_lpips}
\end{table*}

\begin{table*}
\centering
\resizebox{0.7\linewidth}{!}{
\begin{tabular}{l|cccccccc}
Method & \textit{Bicycle} & \textit{Bonsai} & \textit{Counter} & \textit{Garden} & \textit{Kitchen} & \textit{Room} & \textit{Stump} \\
\hline
NeRF~\cite{mildenhall2021nerf} & 0.455 & 0.792 & 0.775 & 0.546 & 0.749 & 0.843 & 0.453\\
mip-NeRF ~\cite{Barron_2021_ICCV} & 0.454 & 0.818 & 0.779 & 0.543 & 0.745 & 0.851 & 0.517\\
mip-NeRF 360~\cite{Barron_2022_CVPR} & 0.685 & 0.941 & 0.894 & 0.813 & 0.920 & 0.913 & 0.744\\
TensoRF~\cite{chen2022tensorf} & 0.345 & 0.754 & 0.681 & 0.411 & 0.552 & 0.777 & 0.387\\
DVGO~\cite{Sun_2022_CVPR} & 0.365 & 0.553 & 0.728 & 0.476 & 0.627 & 0.814 & 0.428\\
EgoNeRF & 0.464 & 0.847 & 0.767 & 0.631 & 0.765 & 0.853 & 0.576\\
\end{tabular}
}
\caption{Per-scene quantitative results in terms of SSIM in mip-NeRF 360~\cite{Barron_2022_CVPR} dataset.}
\label{tab:mip-nerf_ssim}
\end{table*}


\end{document}